\documentclass{article}

\usepackage{microtype}
\usepackage{graphicx}
\usepackage{subcaption}
\usepackage{booktabs} 

\usepackage{hyperref}

\usepackage[preprint]{icml2026}

\usepackage{amsmath}
\usepackage{amssymb}
\usepackage{mathtools}
\usepackage{amsthm}

\usepackage[capitalize,noabbrev]{cleveref}

\theoremstyle{plain}

\theoremstyle{definition}

\theoremstyle{remark}

\usepackage[textsize=tiny]{todonotes}
\usepackage{multirow}
\icmltitlerunning{TRACE: A Temporal Conditional Estimation for Multimodal Time Series Foundation Models}

\begin{document}

\twocolumn[
  \icmltitle{
  TRACE: A Temporal  Conditional Estimation  for Multimodal Time Series Foundation Models
}



  \icmlsetsymbol{equal}{*}

  \begin{icmlauthorlist}
    \icmlauthor{Ziwen Kan}{equal,yyy}
    \icmlauthor{Yishuo Chen}{equal,zzz}
    \icmlauthor{Kecheng Li}{equal}
    \icmlauthor{Andrew Wen}{}
    \icmlauthor{Xiaomeng Wang}{}
    \icmlauthor{Liwei Wang}{}
    \icmlauthor{Jihao Duan}{zzz}
    \icmlauthor{Song Wang}{yyy}
    \icmlauthor{Hongfang Liu}{}
    \icmlauthor{Tianlong Chen}{zzz}
  \end{icmlauthorlist}
  \icmlaffiliation{zzz}{Department of Computer Science, University of North Carolina at Chapel}
  \icmlaffiliation{yyy}{Department of Computer Science, University of Central Florida}


  \icmlcorrespondingauthor{Tianlong Chen}{tianlong@cs.unc.edu}

  \icmlkeywords{Machine Learning, ICML}

  \vskip 0.3in
]



\printAffiliationsAndNotice{\icmlEqualContribution}

\begin{abstract}
Time series foundation models (TS-FMs) aim to learn generalizable temporal representations
that can be adapted to a wide range of downstream tasks.
In real-world multimodal settings, time series are frequently affected by temporal misalignment
and partial modality missingness, where different modalities are observed at heterogeneous
time scales or are partially absent.
Existing approaches typically rely on naive imputation or masking strategies,
which fail to account for cross-modal dependencies and often lead to misaligned
or degraded representations.
We propose \emph{TRACE}, a conditional estimation paradigm for multimodal time series foundation model pipelines
under missingness and irregular sampling,
allowing incomplete target modalities to be systematically inferred from available auxiliary modalities.
We evaluate TRACE on diverse multimodal benchmarks spanning healthcare and affective computing,
including the MIMIC-IV clinical dataset and the CMU-MOSI and CMU-MOSEI benchmarks for multimodal sentiment analysis.
Across a range of downstream prediction tasks and missing-modality settings,
TRACE consistently outperforms prior multimodal fusion approaches,
demonstrating improved robustness to severe modality missingness and more reliable cross-modal representations.
\end{abstract}

 
\section{Introduction}

\begin{figure}[t]
    \centering
    \includegraphics[width=\columnwidth]{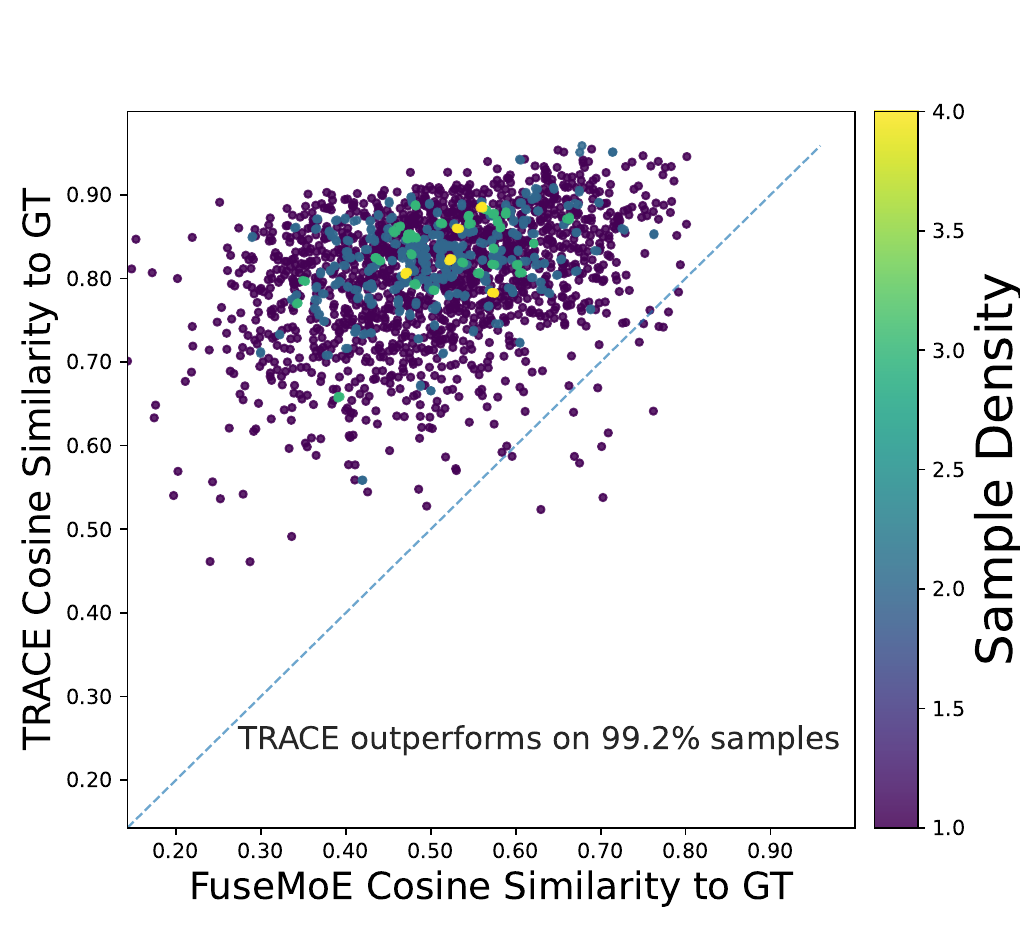}
    \caption{
    Illustration of a multimodal time series setting with a 30\% missing rate, which is common in clinical data.
    We compare sequence-level representations obtained from imputed inputs against
    ground truth (GT) representations derived from fully observed sequences.
    Our paradigm, which treats missing modality inputs as temporal variables to be conditionally estimated from available modalities, outperforms prior value interpolation, yielding internal representations closer to the oracle under the same missingness pattern measured by Cosine Similarity.
    See Appendix~\ref{sec:toy_example} for dataset construction and evaluation details and Appendix~\ref{sec:missing_rate} for additional results.
    }
    \label{fig:motivation}
\end{figure}

Time series analysis~\cite{ahmed2023transformers,liang2024foundation} is a fundamental component of modern data-driven decision making, with applications spanning healthcare~\cite{penfold2013use}, finance~\cite{huang2003applications}, environmental science~\cite{kaur2023autoregressive}, energy systems~\cite{hoffmann2020review}, and autonomous driving~\cite{zhang2022xgboost}. 
Over the past decade, deep learning architectures including CNNs~\cite{ismail2020inceptiontime}, RNNs and LSTMs~\cite{sherstinsky2020fundamentals,yu2019review}, and Transformer-based models~\cite{zhou2021informer,zerveas2021transformer} have been widely adopted for time series modeling.
The emergence of foundation models (FMs)~\cite{bommasani2021opportunities,zhou2024comprehensive} enables reusable, general-purpose representations and unified modeling across tasks and modalities, achieving strong downstream performance in language~\cite{myers2024foundation}, vision~\cite{awais2025foundation}, and graph~\cite{liu2023towards}. 
Motivated by these advances, recent work explores foundation models for time series analysis (TS-FMs) via large-scale pretraining to improve cross-task and cross-domain generalization.
Because time series representations are abstracted from raw temporal processes~\cite{liu2025empowering}, complementary modalities often provide essential contextual signals, motivating growing interest in multimodal TS-FMs.


Despite increasing interest in multimodal TS-FMs, real-world multimodal time series, particularly in healthcare, often exhibit substantial temporal misalignment and modality-level missingness~\cite{liang2024foundation,kottapalli2025foundation,liu2025empowering}. 
These challenges arise from heterogeneous acquisition processes, where some modalities are sampled at fixed rates while others are event-driven, leading to sparse and partially observed sequences.

While this issue has been acknowledged, most existing TS-FMs address missing observations and irregular sampling in a limited or decoupled manner~\cite{cao2024timedit,zhang2023improving}. 
FuseMoE~\cite{han2024fusemoe} represents an important step toward fleximodal modeling by explicitly handling modality-level missingness and temporal misalignment via alignment-based mechanisms such as mTAND~\cite{shukla2021multi}. 
However, under severe sparsity, alignment-based modeling becomes constrained, and FuseMoE resorts to nearest-value filling to synthesize missing observations.
Such deterministic completion collapses conditional uncertainty, potentially distorting temporal dynamics~\cite{fons2025lscd} and biasing downstream fusion.
More broadly, this reflects a paradigm-level limitation shared by many multimodal TS-FMs: 
missing and irregular sampling are often treated as values to be filled rather than latent temporal variables to be conditionally estimated.


To tackle this emerging issue, we propose \emph{TRACE}, a conditional estimation paradigm for multimodal TS-FM pipelines that reformulates how sparse and event-driven observations are handled prior to multimodal fusion. 
Instead of deterministically completing missing observations, TRACE performs temporal conditional estimation by leveraging complementary multimodal context.
Figure~\ref{fig:motivation} illustrates this using a motivational example, showing that conditional estimation yields sequence-level representations that remain closer to the oracle as missingness increases.

Our contributions can be summarized as follows:

\begin{itemize}
\item We identify incomplete and irregular observations as a critical bottleneck in TS-FM pipelines, where deterministic completion is commonly applied prior to fusion.
\item We propose \emph{TRACE} (\underline{T}empo\underline{ra}l \underline{C}onditional \underline{E}stimation), a conditional estimation paradigm for multimodal TS-FM pipelines, instantiated via a diffusion-based mechanism for probabilistic signal-level estimation.
\item Through extensive experiments on diverse multimodal benchmarks, we show that TRACE consistently improves downstream performance across datasets and missingness patterns, establishing temporal conditional estimation as an effective system-level leverage point.
\end{itemize}

\section{Related Work}

\subsection{Multimodal Time Series Foundation Models}

Most existing multimodal TS-FMs implicitly assume that modality-specific representations are fully observed and readily alignable for fusion, limiting their applicability under missingness and irregular sampling.
For example, TimesFM~\cite{das2024decoder}, a decoder-only forecasting FM, requires filtering missing values during preprocessing, restricting its use to curated settings with complete observations. 
PromptCast~\cite{xue2023promptcast} incorporates language signals for forecasting but similarly assumes complete observations prior to fusion.
Other approaches rely on task-specific interpolation; for instance, TEMPO~\cite{cao2023tempo} applies zero-filling in company investment analysis, which depends on domain knowledge and does not generalize well across multimodal settings.
DAM~\cite{darlow2024dam} models time series as continuous functions of time, naturally supporting irregular observations and horizon-free forecasting.
However, it primarily targets forecasting and signal-level regression, and does not address representation-level alignment and fusion across heterogeneous modalities.

More recent TS-FMs begin to explicitly consider missingness.
FlexMoE~\cite{yun2024flex} improves robustness through flexible expert routing, yet still assumes that modality representations are available prior to fusion.
TimeDiT~\cite{cao2024timedit} addresses missingness via masking and diffusion-based denoising, but operates exclusively in unimodal settings and does not incorporate cross-modal conditioning.
FuseMoE~\cite{han2024fusemoe} further extends this line to modality-level missingness through mixture-of-experts fusion; however, it relies on deterministic imputation to populate missing modality representations before fusion, treating estimation as a preprocessing step rather than a learned, conditional process.
Overall, existing TS-FMs either ignore missingness or rely on deterministic or unimodal estimation, without performing cross-modal conditional estimation prior to multimodal fusion.

\subsection{Time Series Imputation}

Time series imputation has been studied primarily in unimodal settings \cite{wang2024deep}. 
Earlier approaches include predictive or reconstruction-based models such as M-RNN \cite{yoon2018estimating}, TimesNet \cite{wu2022timesnet}, and MVI \cite{bansal2021missing}, while generative variants include GP-VAE \cite{fortuin2020gp} and GRUI-GAN \cite{luo2018multivariate}. 
Recent diffusion methods, including CSDI \cite{tashiro2021csdi}, SSSD \cite{alcaraz2022diffusion}, and LSCD \cite{fons2025lscd}, further improve conditional signal completion, and foundation models such as Timer \cite{liu2024timer} and MOMENT \cite{goswami2024moment} can be adapted to masked imputation objectives. 
However, these methods still focus on recovering missing values within a single time series and operate mainly in raw signal space, rather than performing cross-modal conditional estimation to preserve coherence before multimodal fusion.

In contrast to existing multimodal time series foundation models,
we propose a TS-FM paradigm that explicitly integrates missing-modality
estimation into the multimodal modeling pipeline prior to fusion.
While preserving flexible multimodal fusion,
our approach addresses modality-level missingness through conditional estimation,
leveraging available modalities to generate more faithful and context-aware
representations for partially observed ones.

\section{Method} 
\begin{figure*}[thbp]
    \centering
    \includegraphics[width=\linewidth]{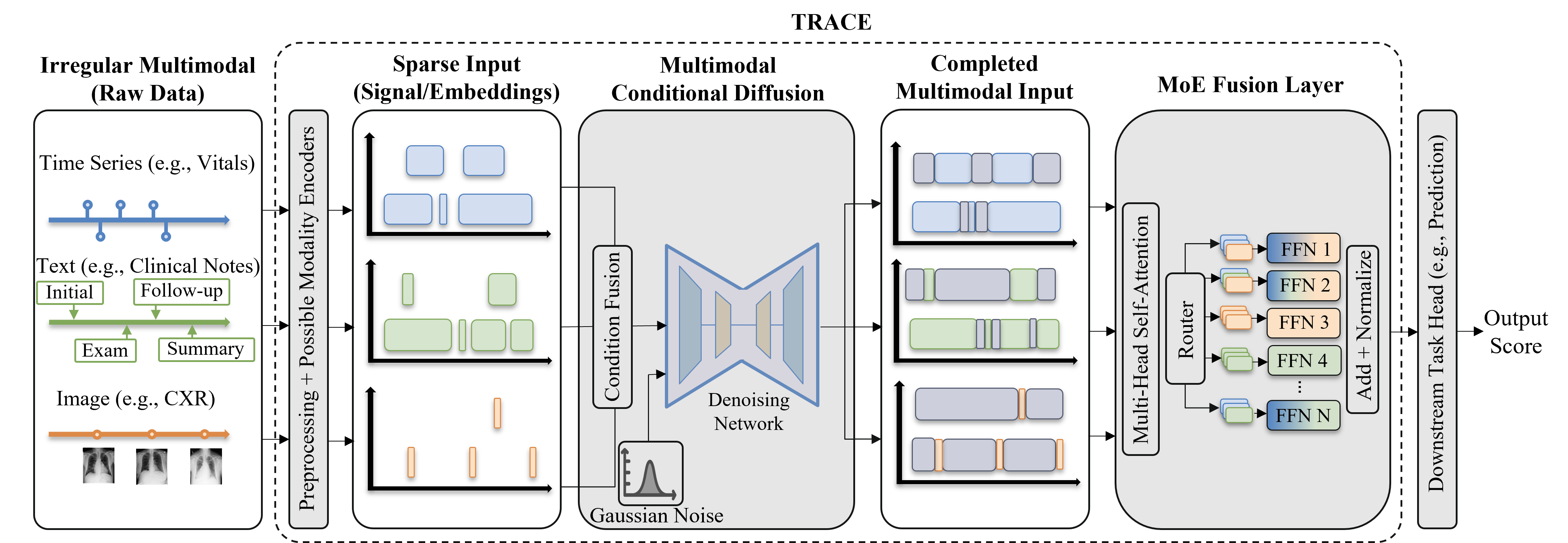}
    \caption{
    Overall architecture of TRACE, a two-stage conditional estimation paradigm for multimodal time series foundation model pipelines.
    Given incomplete and irregular multimodal inputs, TRACE first performs \emph{multimodal conditional diffusion},
    where each target modality is conditionally completed at the representation level by leveraging its observed components
    and an MoE-gated cross-modal context constructed from available auxiliary modalities.
    The resulting completed multimodal representations are then fed into a \emph{Mixture-of-Experts (MoE) fusion layer}
    to produce a unified embedding for downstream task prediction.
    }
    \label{fig:overall}
\end{figure*}

\begin{figure}[t]
    \centering
    \includegraphics[width=\linewidth]{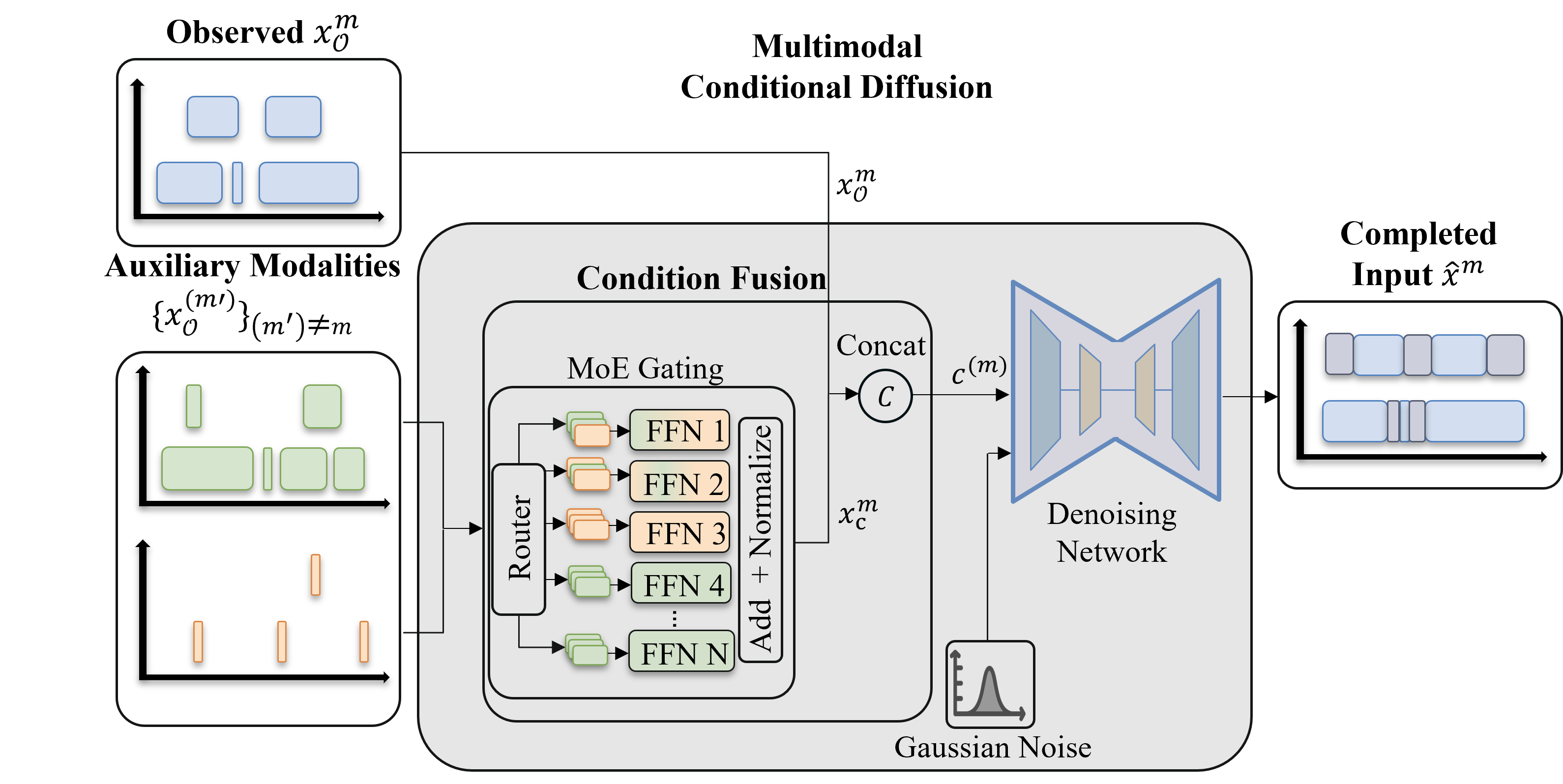}
    \caption{
    Illustration of the multimodal conditional diffusion block.
    It performs denoising on a partially observed target modality by injecting Gaussian noise into missing components and leveraging selected auxiliary modality representations as conditioning signals.
    }
    \label{fig:trace}
\end{figure}

\subsection{Problem Statement}

We study multimodal time series data with heterogeneous modalities and irregular observations.
Let $\mathcal{M}=\{1,\dots,M\}$ denote the set of modalities.
For each modality $m\in\mathcal{M}$, we denote its input as a modality-specific representation,
which may be partially observed and misaligned across modalities.
Depending on the dataset and modality, the raw inputs may undergo appropriate preprocessing and encoding.
After preprocessing, we denote the input for modality $m$ as
$x^{(m)} \in \mathbb{R}^{d_m}$, where $d_m$ denotes the dimensionality of the modality $m$. Note that such preprocessing operates only on observed components and does not resolve missingness; consequently, $x^{(m)}$ remains partially observed.
For each modality $m$, we consider partially observed representations and introduce
a modality-specific binary observation mask $r^{(m)} \in \{0,1\}^{L \times d_m}$,
where $L$ denotes the temporal length.
Using $r^{(m)}$, we write the observed and missing components of the representation
as $x^{(m)}_{\mathcal{O}} = r^{(m)} \odot x^{(m)}$ and
$x^{(m)}_{\mathcal{U}} = (1 - r^{(m)}) \odot x^{(m)}$, respectively.

\subsection{Overall Architecture}

As illustrated in Figure~\ref{fig:overall}, TRACE defines a complete modeling paradigm for multimodal time series foundation models.
Under TRACE, multimodal modeling is structured as a two-stage pipeline consisting of multimodal conditional diffusion followed by multimodal fusion and prediction.

In the first stage, TRACE performs \emph{multimodal conditional diffusion}, which treats temporal conditional estimation as an intermediate objective prior to fusion. For each modality $m\in\mathcal{M}$, the model conditions on the observed components of $x^{(m)}_{\mathcal{O}}$ together with an aggregated cross-modal context derived from the remaining modalities $x_c^{(m)}$, and probabilistically estimates the missing or unobserved components $x^{(m)}_{\mathcal{U}}$.
This procedure can be instantiated across modalities to produce conditionally estimated modality-specific representations.
Formally, for each target modality $m$, TRACE aims to model the conditional distribution
$p_\theta\!\left(x^{(m)}_{\mathcal{U}} \mid x^{(m)}_{\mathcal{O}}, x_c^{(m)}\right)$
of the unobserved components given the observed part and an aggregated cross-modal context
$x_c^{(m)}$ constructed from the remaining modalities $\mathcal{M}\setminus\{m\}$,
where $\theta$ denotes the parameters of the conditional diffusion model.
This conditional distribution is instantiated via a conditional diffusion model,
from which completed modality representations are obtained for downstream processing.

In the second stage, TRACE performs multimodal fusion and downstream prediction.
Specifically, we adopt the MoE fusion layer from FuseMoE \cite{han2024fusemoe} to aggregate the completed modality-specific representations and produce a unified fused embedding, which is subsequently consumed by task-specific prediction heads.

\subsection{Multimodal Conditional Diffusion}

We now describe how TRACE instantiates the first stage of the paradigm, i.e., multimodal conditional diffusion.

\paragraph{MoE-based Condition Fusion.}
To robustly construct the cross-modal conditioning signal, we aggregate observed components from auxiliary modalities $\mathcal{M}\setminus\{m\}$ using an MoE-inspired gating mechanism, yielding the cross-modal context $x_c^{(m)}$.
Specifically, a gating function assigns a scalar importance score to each auxiliary modality.
The gating logits are computed as inner product
$ h^{(m')} = \langle w, x^{(m')}_\mathcal{O} \rangle $
where $w \in \mathbb{R}^{d_{m'}}$ is a learnable parameter and $d_{m'}$ denotes the dimensionality of the auxiliary modality. 
The weights are then normalized across available modalities using a softmax function.

The gating weights are obtained by normalizing the logits across auxiliary modalities
using a softmax function:
\begin{equation}
\label{eq:gate_weights}
\pi^{(m')} =
\frac{\exp\!\left(h^{(m')}\right)}
{\sum\limits_{j \in \mathcal{M}\setminus\{m\}} \exp\!\left(h^{(j)}\right)},
\end{equation}
Using gating weights, the cross-modal context for the target modality $m$ is constructed as a weighted aggregation of auxiliary modality representations:
\begin{equation}
\label{eq:cross_modal_context}
x_c^{(m)} =
\sum_{m' \in \mathcal{M}\setminus\{m\} }
\pi^{(m')} \, x^{(m')}_\mathcal{O}.
\end{equation}
This MoE-based gating mechanism allows TRACE to emphasize informative auxiliary modalities while suppressing noisy or unreliable ones, resulting in a robust cross-modal context for conditional estimation.

Finally, the conditioning variable $c^{(m)}$ is constructed by concatenating
the observed part of the target modality $x^{(m)}_{\mathcal{O}}$
with the MoE-aggregated cross-modal context $x^{(m)}_c$:
\begin{equation}
c^{(m)} = \mathrm{Concat}\!\left(x^{(m)}_{\mathcal{O}},\, x^{(m)}_c\right).
\end{equation}



\paragraph{Conditional Diffusion.}
We instantiate the intermediate conditional estimation objective using a conditional denoising diffusion model.
Following DDPM~\cite{ho2020denoising}, we adopt the noise-prediction parameterization
and model conditional estimation through a Markovian diffusion process consisting of
a forward noising process and a conditional reverse denoising process.
To distinguish diffusion latents from modality notation, we denote by $z^{(m)}_t$
the noisy latent variable of modality $m$ at diffusion step $t$, which is defined
only over the unobserved components $x^{(m)}_{\mathcal{U}} = (1 - r^{(m)}) \odot x^{(m)}$, while observed entries are kept fixed and used as conditioning signals throughout the diffusion
process:
\begin{equation}
q\!\left(
z^{(m)}_{t} \mid z^{(m)}_{t-1}
\right)
=
\mathcal{N}\!\left(
z^{(m)}_{t};
\sqrt{1-\beta_t}\,z^{(m)}_{t-1},\,
\beta_t I
\right),
\end{equation}
where $\{\beta_t\}_{t=1}^T$ denotes a predefined noise schedule with $\beta_t \in (0,1)$,
and $I$ is the identity covariance matrix.

Equivalently, the diffusion trajectory admits the standard closed-form reparameterization
$
z^{(m)}_t
=
\sqrt{\bar{\alpha}_t}\,x^{(m)}_0
+
\sqrt{1-\bar{\alpha}_t}\,\epsilon,
$
with $\epsilon \sim \mathcal{N}(0,I)$.

For a target modality $m$, we define a conditional generative process that models
the distribution of missing components conditioned on multimodal context as
\begin{equation}
\label{eq:conditional_diffusion_process}
p_\theta\!\left(
z^{(m)}_{0:T} \mid c^{(m)}
\right)
:=
p\!\left(z^{(m)}_T\right)
\prod_{t=1}^{T}
p_\theta\!\left(
z^{(m)}_{t-1} \mid z^{(m)}_t,\ c^{(m)}
\right)
\end{equation}
where $z^{(m)}_T \sim \mathcal{N}(0, I)$ is sampled from the standard Gaussian prior,
and $z^{(m)}_0$ denotes the completed target-modality representation, where the observed entries are fixed and the missing entries are generated by the reverse process.
The conditional reverse transition is defined as
\begin{equation}
\label{eq:conditional_reverse}
p_\theta\!\left(
z^{(m)}_{t-1} \mid z^{(m)}_t,\ c^{(m)}
\right)
=
\mathcal{N}\!\left(
z^{(m)}_{t-1};
\mu_\theta,\,
\sigma_t^2 I
\right),
\end{equation}
where $c^{(m)}$ denotes the multimodal conditioning context.
The mean term is parameterized as
$\mu_\theta \equiv \mu_\theta\!\left(z^{(m)}_t, t \mid c^{(m)}\right)$,
following the standard DDPM reverse-process formulation with
$\epsilon_\theta\!\left(z^{(m)}_t, t \mid c^{(m)}\right)$
as a conditional denoising network.
The variance $\sigma_t^2$ is a time-dependent scalar determined by the predefined noise schedule.
This formulation enables TRACE to incorporate multimodal context into the denoising process
while preserving the probabilistic structure and training objective of conditional DDPM.


\paragraph{MoE Fusion Layer.}
After obtaining modality representations from the preceding conditional diffusion step,
TRACE performs multimodal fusion using the MoE fusion layer from FuseMoE.
We adopt this fusion module without architectural modification.
Specifically, the FuseMoE fusion layer consists of a multi-head self-attention block,
followed by a sparse Mixture-of-Experts (MoE) gating mechanism that routes representations
to a shared pool of feed-forward experts, and a residual \texttt{Add\,+\,Norm} operation
to aggregate expert outputs into a unified fused representation for downstream prediction.
FuseMoE provides several router design variants; unless otherwise specified, we use its
default \emph{joint-experts} configuration, where modality-aware routing weights are
computed jointly over all modalities and experts, which is well-suited for aggregating
heterogeneous modality representations in multimodal settings.

\subsection{Training of TRACE}

TRACE is trained in a two-stage manner, corresponding to multimodal conditional diffusion and MoE-based multimodal fusion.

\paragraph{Training of Multimodal Conditional Diffusion}
The denoising network $\epsilon_\theta$ is trained according to the standard diffusion objective:
\begin{equation}
\label{eq:conditional_loss}
\mathcal{L}_{\text{diff}}(\theta)
=
\mathbb{E}_{z^{(m)}_0,\,\epsilon,\,t}
\left[
\left\|
\epsilon -
\epsilon_\theta\!\left(
z^{(m)}_t,\ t \mid c^{(m)}
\right)
\right\|_2^2
\right],
\end{equation}
where $\epsilon \sim \mathcal{N}(0, I)$ and
$z^{(m)}_t = \sqrt{\bar{\alpha}_t}\,z^{(m)}_0 + \sqrt{1-\bar{\alpha}_t}\,\epsilon$
is obtained via the standard DDPM reparameterization.

As ground-truth missing values are unavailable during training, we adopt the self-supervised masking strategy proposed in CSDI~\cite{tashiro2021csdi}.
Specifically, a subset of observed values is randomly masked and treated as imputation targets, while the remaining observations serve as conditioning signals.
The model is trained to denoise the masked targets under the diffusion process, conditioned on unmasked observations and cross-modal context. While the masking mechanism follows CSDI, the conditioning information in our setting is inherently multimodal, enabling cross-modal conditional estimation with richer information.

\paragraph{Training of MoE Fusion Layer.}
The MoE fusion layer is trained jointly with downstream prediction heads using task-specific supervision.
Following FuseMoE, we incorporate an entropy-based regularization term as a load-balancing loss~\cite{shazeer2017outrageously}
to encourage balanced and stable expert utilization across modalities,
mitigating expert collapse where routing decisions concentrate on a small subset of experts.
Beyond enforcing uniform expert usage, this regularization also implicitly encourages
modality-specific input embeddings to produce more confident and discriminative routing distributions,
thereby reducing uncertainty in expert selection.
Specifically, given $M$ modalities, the entropy regularization loss is defined as:
\begin{equation}
\mathcal{L}_{\text{MoE}}
=
\frac{1}{M} \sum_{j=1}^{M} \mathcal{H}(\hat{p}_{m_j}(E))
-
\mathcal{H}\!\left(
\frac{1}{M} \sum_{j=1}^{M} \hat{p}_{m_j}(E)
\right),
\end{equation}
where $\mathcal{H}$ is the entropy operator and $\hat{p}_{m_j}(E)$ denotes the routing distribution over experts
for the $j$-th modality, approximated by averaging routing probabilities across observations of that modality.
This regularization term is added to the task-specific loss to form the overall training objective.

\begin{table*}[thbp]
\centering
\small
\setlength{\tabcolsep}{3.5pt}
\renewcommand{\arraystretch}{1.05}
\caption{
Performance comparison on CMU-MOSI and CMU-MOSEI datasets.
MAE denotes mean absolute error for sentiment regression,
Acc-2 and F1 correspond to binary sentiment classification accuracy and F1 score,
and Corr indicates the Pearson correlation between predicted and ground-truth sentiment scores.
The best results are highlighted in \textbf{bold font}, and the second-best results are \underline{underlined}.
Results are reported as mean $\pm$ standard deviation over 3 runs.
}
\begin{tabular}{l|cccc|cccc}
\toprule
\textbf{Method $\backslash$ Task}
& \multicolumn{4}{c|}{\textbf{MOSI Dataset}}
& \multicolumn{4}{c}{\textbf{MOSEI Dataset}} \\
\cmidrule(lr){2-9}
& MAE$\downarrow$ & Acc-2$\uparrow$ & Corr$\uparrow$ & F1$\uparrow$
& MAE$\downarrow$ & Acc-2$\uparrow$ & Corr$\uparrow$ & F1$\uparrow$ \\
\midrule

\textbf{MulT}
& 0.86 $\pm$ 0.01 & 84.10 $\pm$ 0.21 & 0.71 $\pm$ 0.02 & 83.90 $\pm$ 0.27
& 0.58 $\pm$ 0.02 & 82.51 $\pm$ 0.41 & 0.71 $\pm$ 0.04 & 82.31 $\pm$ 0.27 \\

\textbf{MAG}
& \underline{0.71 $\pm$ 0.04} & 86.10 $\pm$ 0.44 & 0.80 $\pm$ 0.03 & 86.00 $\pm$ 0.59
& \underline{0.57 $\pm$ 0.07} & 85.56 $\pm$ 0.22 & \textbf{0.79 $\pm$ 0.02} & 84.50 $\pm$ 0.18 \\

\textbf{TFN}
& 0.90 $\pm$ 0.02 & 80.81 $\pm$ 0.34 & 0.70 $\pm$ 0.04 & 80.70 $\pm$ 0.18
& 0.59 $\pm$ 0.03 & 82.50 $\pm$ 0.58 & 0.68 $\pm$ 0.02 & 82.10 $\pm$ 0.41 \\

\textbf{FuseMoE}
& 0.72 $\pm$ 0.02 & \underline{86.54 $\pm$ 0.58} & \underline{0.80 $\pm$ 0.01} & \underline{86.45 $\pm$ 0.50}
& \textbf{0.53 $\pm$ 0.01} & \underline{86.10 $\pm$ 0.19} & 0.78 $\pm$ 0.01 & \underline{86.07 $\pm$ 0.27} \\

\textbf{TRACE (ours)}
& \textbf{0.70 $\pm$ 0.01} & \textbf{87.09 $\pm$ 0.58} & \textbf{0.81 $\pm$ 0.01} & \textbf{87.04 $\pm$ 0.56}
& \textbf{0.53 $\pm$ 0.01} & \textbf{86.44 $\pm$ 0.26} & \underline{0.78 $\pm$ 0.00} & \textbf{86.45 $\pm$ 0.21} \\

\bottomrule
\end{tabular}

\label{tab:mosi_mosei}
\end{table*}

\begin{table*}[thbp]
\centering
\small
\setlength{\tabcolsep}{3.5pt}
\renewcommand{\arraystretch}{1.05}
\caption{
Performance comparison of TRACE and baseline methods on vital-sign and clinical-note tasks on the MIMIC-IV dataset.
AUROC and F1 are used to evaluate prediction and classification performance.
The best results are highlighted in \textbf{bold font}, and the second-best results are \underline{underlined}.
Results are reported as mean $\pm$ standard deviation over 3 runs, with the overall average across tasks.
}
\begin{tabular}{l|cc|cc|cc|cc}
\toprule
\textbf{Method $\backslash$ Task}
& \multicolumn{2}{c|}{\textbf{48-IHM}}
& \multicolumn{2}{c|}{\textbf{LOS}}
& \multicolumn{2}{c|}{\textbf{25-PHE}}
& \multicolumn{2}{c}{\textbf{Average}} \\
\cmidrule(lr){2-7} \cmidrule(lr){8-9}
& AUROC$\uparrow$ & F1$\uparrow$
& AUROC$\uparrow$ & F1$\uparrow$
& AUROC$\uparrow$ & F1$\uparrow$
& AUROC$\uparrow$ & F1$\uparrow$ \\
\midrule

\textbf{MISTS}
& 75.06 $\pm$ 1.03 & 45.61 $\pm$ 0.34
& 80.56 $\pm$ 0.33 & 73.01 $\pm$ 0.52
& 69.45 $\pm$ 0.72 & 28.59 $\pm$ 0.46
& 75.02 & 49.07 \\

\textbf{MulT}
& 75.95 $\pm$ 0.84 & 38.81 $\pm$ 0.22
& 81.36 $\pm$ 1.32 & 73.45 $\pm$ 0.59
& 66.58 $\pm$ 0.41 & 28.55 $\pm$ 0.31
& 74.63 & 46.94 \\

\textbf{MAG}
& 75.82 $\pm$ 0.73 & 42.55 $\pm$ 0.82
& 81.13 $\pm$ 0.66 & 72.51 $\pm$ 0.27
& 69.55 $\pm$ 0.67 & 27.86 $\pm$ 0.29
& 75.50 & 47.64 \\

\textbf{TFN}
& 78.76 $\pm$ 0.79 & 40.61 $\pm$ 0.41
& 80.71 $\pm$ 0.45 & 73.84 $\pm$ 0.61
& 69.18 $\pm$ 0.32 & 28.52 $\pm$ 0.22
& 76.22 & 47.66 \\

\textbf{HAIM}
& 79.65 $\pm$ 0.00 & 39.79 $\pm$ 0.00
& \textbf{82.58 $\pm$ 0.00} & 73.81 $\pm$ 0.00
& 63.39 $\pm$ 0.00 & \textbf{42.13 $\pm$ 0.00}
& 75.21 & \underline{51.91} \\

\textbf{MAESTRO}
& 81.86 $\pm$ 0.07 & 35.07 $\pm$ 2.28
& 79.50 $\pm$ 0.09 & 73.61 $\pm$ 0.39
& \underline{74.18 $\pm$ 0.46} & 30.55 $\pm$ 0.44
& 78.51 & 46.41 \\

\textbf{FuseMoE}
& \underline{82.83 $\pm$ 0.35} & \underline{46.93 $\pm$ 0.68}
& 80.86 $\pm$ 1.42 & \underline{75.08 $\pm$ 0.60}
& \underline{73.80 $\pm$ 1.72} & 28.43 $\pm$ 3.05
& \underline{79.16} & 50.15 \\

\textbf{TRACE (ours)}
& \textbf{83.29 $\pm$ 0.12} & \textbf{47.07 $\pm$ 0.63}
& \underline{81.44 $\pm$ 0.10} & \textbf{77.60 $\pm$ 0.37}
& \textbf{75.00 $\pm$ 0.54} & \underline{33.13 $\pm$ 1.78}
& \textbf{79.91} & \textbf{52.60} \\

\bottomrule
\end{tabular}

\label{tab:MIMIC}
\end{table*}

\section{Experiments}
\subsection{Datasets and Settings}

We evaluate TRACE on CMU-MOSI, CMU-MOSEI~\cite{zadeh2016multimodal}, and MIMIC-IV~\cite{johnson2020mimic}, spanning sentiment analysis and clinical prediction.
CMU-MOSI and CMU-MOSEI are multimodal sentiment datasets with aligned text, audio, and visual modalities, containing 1,284 / 229 / 686 and 16,326 / 1,871 / 4,659 train/validation/test samples, respectively.
Following prior work~\cite{han2021improving,han2024fusemoe}, we report MAE, Pearson correlation (Corr), accuracy (Acc), and F1 on these datasets.
MIMIC-IV includes four modalities: vital signs (Vital), clinical notes (Notes), chest X-rays (CXR), and electrocardiograms (ECG).
We consider 48-hour in-hospital mortality (48-IHM), length-of-stay (LOS), and 25-type phenotype classification (25-PHE), using standard 70/15/15 train/validation/test splits with 35,129 ICU stays for 48-IHM and LOS and 73,173 for 25-PHE; evaluation uses AUROC and F1 score~\cite{zhang2023improving,lin2019predicting,arbabi2019identifying}.
Unless otherwise specified, we adopt the same experimental protocol introduced in FuseMoE to ensure fair comparison. Additional implementation details are provided in the Appendix.

Following FuseMoE~\cite{han2024fusemoe}, we compare TRACE against representative multimodal baselines, including MISTS~\cite{zhang2023improving}, TFN~\cite{zadeh2017tensor}, MAG~\cite{rahman2020integrating}, MAESTRO~\cite{mohapatra2025maestro}, and FuseMoE~\cite{han2024fusemoe}.
To isolate the contribution of the estimation module, we further compare TRACE against CSDI~\cite{tashiro2021csdi} and SSSD~\cite{alcaraz2022diffusion} in a separate estimation-oriented comparison under the same multimodal protocol.

\begin{table*}[thbp]
\centering
\small
\setlength{\tabcolsep}{5pt}
\renewcommand{\arraystretch}{1.12}
\caption{Performance comparison of TRACE and baseline methods on MIMIC-IV with different modality combinations. Results are reported as mean $\pm$ standard deviation over 3 runs, with the overall average across tasks.}
\begin{tabular}{l|cc|cc|cc|cc}
\toprule
\multicolumn{9}{c}{\textbf{Vital \& Notes \& CXR}} \\
\midrule
\textbf{Method $\backslash$ Task}
& \multicolumn{2}{c|}{\textbf{48-IHM}}
& \multicolumn{2}{c|}{\textbf{LOS}}
& \multicolumn{2}{c|}{\textbf{25-PHE}}
& \textbf{Avg} & \textbf{Avg} \\
\cline{2-9}
& AUROC$\uparrow$ & F1$\uparrow$
& AUROC$\uparrow$ & F1$\uparrow$
& AUROC$\uparrow$ & F1$\uparrow$
& AUROC$\uparrow$ & F1$\uparrow$ \\
\midrule

\textbf{HAIM}
& 78.87 $\pm$ 0.00 & 39.78 $\pm$ 0.00
& \textbf{82.46 $\pm$ 0.00} & 72.75 $\pm$ 0.00
& 63.57 $\pm$ 0.00 & \textbf{42.80 $\pm$ 0.00}
& 74.97 & \underline{51.78} \\

\textbf{FuseMoE}
& \underline{81.33 $\pm$ 0.08} & \underline{43.19 $\pm$ 1.52}
& 79.79 $\pm$ 1.34 & \underline{74.84 $\pm$ 0.38}
& \underline{75.54 $\pm$ 1.10} & 31.93 $\pm$ 2.62
& \underline{78.89} & 49.99 \\

\textbf{TRACE (ours)}
& \textbf{83.60 $\pm$ 0.22} & \textbf{48.99 $\pm$ 0.23}
& \underline{82.12 $\pm$ 0.06} & \textbf{75.54 $\pm$ 0.68}
& \textbf{75.94 $\pm$ 1.24} & \underline{34.93 $\pm$ 2.90}
& \textbf{80.55} & \textbf{53.15} \\

\midrule
\multicolumn{9}{c}{\textbf{Vital \& Notes \& CXR \& ECG}} \\
\midrule

\textbf{HAIM}
& 78.87 $\pm$ 0.00 & 39.78 $\pm$ 0.00
& \textbf{82.46 $\pm$ 0.00} & 72.75 $\pm$ 0.00
& 63.82 $\pm$ 0.00 & \textbf{43.20 $\pm$ 0.00}
& 75.05 & \underline{51.91} \\

\textbf{FuseMoE}
& \underline{81.09 $\pm$ 0.24} & \textbf{44.05 $\pm$ 0.42}
& 80.04 $\pm$ 1.24 & \underline{74.78 $\pm$ 0.18}
& \underline{74.76 $\pm$ 0.33} & 32.63 $\pm$ 2.23
& \underline{78.63} & 50.49 \\

\textbf{TRACE (ours)}
& \textbf{81.68 $\pm$ 0.29} & \underline{43.57 $\pm$ 0.38}
& \underline{81.21 $\pm$ 0.59} & \textbf{75.50 $\pm$ 0.32}
& \textbf{76.14 $\pm$ 0.43} & \underline{38.94 $\pm$ 2.25}
& \textbf{79.67} & \textbf{52.67} \\

\bottomrule
\end{tabular}

\label{tab:MIMIC_3v4}
\end{table*}

\paragraph{Comparison with Diffusion-Based Imputation Baselines.}
To isolate the effect of the estimation module, we compare TRACE against CSDI~\cite{tashiro2021csdi} and SSSD~\cite{alcaraz2022diffusion} under the same downstream multimodal fusion pipeline on the 48-IHM task.

\begin{table*}[t]
\centering
\small
\setlength{\tabcolsep}{4pt}
\renewcommand{\arraystretch}{1.10}
\caption{
Estimation-oriented comparison on the 48-IHM task of MIMIC-IV under identical multimodal settings.
SSSD and CSDI are diffusion-based unimodal imputation baselines, while TRACE performs cross-modal conditional estimation before fusion.
}
\begin{tabular}{l|cc|cc|cc}
\toprule
\textbf{Method}
& \multicolumn{2}{c|}{\textbf{Vital \& Notes}}
& \multicolumn{2}{c|}{\textbf{Vital \& Notes \& CXR}}
& \multicolumn{2}{c}{\textbf{Vital \& Notes \& CXR \& ECG}} \\
\cmidrule(lr){2-3} \cmidrule(lr){4-5} \cmidrule(lr){6-7}
& AUROC$\uparrow$ & F1$\uparrow$
& AUROC$\uparrow$ & F1$\uparrow$
& AUROC$\uparrow$ & F1$\uparrow$ \\
\midrule
\textbf{SSSD}
& 81.07 $\pm$ 0.86 & 43.30 $\pm$ 1.49
& 82.20 $\pm$ 0.09 & \underline{47.13 $\pm$ 0.63}
& 81.13 $\pm$ 1.30 & \underline{43.08 $\pm$ 2.39} \\

\textbf{CSDI}
& \underline{83.15 $\pm$ 0.08} & \textbf{47.18 $\pm$ 0.41}
& \underline{82.31 $\pm$ 0.64} & 44.23 $\pm$ 4.03
& \underline{81.47 $\pm$ 1.18} & 42.58 $\pm$ 2.30 \\

\textbf{TRACE (ours)}
& \textbf{83.29 $\pm$ 0.12} & \underline{47.07 $\pm$ 0.63}
& \textbf{83.60 $\pm$ 0.22} & \textbf{48.99 $\pm$ 0.23}
& \textbf{81.68 $\pm$ 0.29} & \textbf{43.57 $\pm$ 0.38} \\
\bottomrule
\end{tabular}
\label{tab:estimation_baselines}
\end{table*}










\subsection{Main Results}

\paragraph{CMU-MOSI and MOSEI Datasets.}
These benchmarks involve aligned visual, acoustic, and textual modalities for
multimodal sentiment and emotion analysis. The results on CMU-MOSI and CMU-MOSEI are reported in Table~\ref{tab:mosi_mosei}.
For fair comparison with prior work, we adopt the same feature extraction
pipeline as FuseMoE, using a pre-trained T5~\cite{raffel2020exploring} encoder for
text, librosa~\cite{mcfee2015librosa} for acoustic features, and EfficientNet~\cite{tan2019efficientnet}
for visual representations. 

On CMU-MOSI, TRACE consistently outperforms all baseline methods across all evaluation metrics, including MAE, accuracy, correlation, and F1 score.
Compared to FuseMoE, TRACE achieves lower prediction error and higher
classification and correlation performance.
On CMU-MOSEI, TRACE achieves the best overall performance across all evaluation metrics.
In particular, TRACE attains the lowest MAE and the highest accuracy and F1 score,
while achieving competitive correlation performance compared to prior methods.
Compared to FuseMoE, TRACE consistently improves classification accuracy and F1 score
and maintains comparable correlation with substantially reduced variance,
indicating more stable and effective cross-modal estimation under missing-modality settings.

\paragraph{MIMIC-IV Dataset.}
Following prior work, we adopt HAIM~\cite{soenksen2022integrated}, a multimodal clinical data processing pipeline specifically tailored for MIMIC-IV.
The results on MIMIC-IV are summarized in Table~\ref{tab:MIMIC}.
Overall, TRACE achieves the strongest average performance across tasks, outperforming all multimodal baselines in terms of both AUROC and F1 when averaged over 48-IHM, LOS, and 25-PHE.
This indicates that conditional cross-modal estimation provides consistent benefits under heterogeneous clinical objectives and missingness patterns.
On the 48-IHM task, TRACE attains the best performance across both metrics, exceeding all baseline multimodal methods.
For LOS and phenotype prediction, TRACE consistently improves over FuseMoE,
demonstrating the effectiveness of modeling cross-modal dependencies beyond heuristic imputation.
While HAIM benefits from domain-specific inductive biases for physiological time series aggregation,
TRACE achieves strong and competitive results across tasks without task-specific design.
We further isolate the estimation module in Table~\ref{tab:estimation_baselines} by comparing TRACE against CSDI and SSSD under the same downstream fusion pipeline.
In the two-modality setting, CSDI remains competitive in AUROC, but TRACE delivers more consistent gains in F1 as the modality count increases.
This emphasis on F1 is important because our goal is not only to preserve ranking quality, but also to recover clinically useful positive evidence under partial observation so that downstream fusion yields better precision-recall trade-offs.
The stronger F1 under richer modality combinations suggests that explicit cross-modal conditional estimation becomes increasingly beneficial when clinical evidence is distributed across heterogeneous and only partially observed sources.

We note that, except for HAIM, most baseline methods were not originally designed to be agnostic
to the quantity and diversity of input modalities.
As a result, several methods cannot be directly extended to settings with additional and partially
observed modalities without non-trivial architectural or routing modifications,
and are therefore omitted from Table~\ref{tab:MIMIC_3v4}.
Overall, TRACE achieves the strongest average performance across modality settings,
consistently outperforming FuseMoE and remaining competitive with HAIM as additional modalities are introduced.
This demonstrates that TRACE maintains robust multimodal representations under increasingly complex fusion scenarios.
Compared to the two-modality setting, FuseMoE benefits from the inclusion of CXR in several cases, indicating that complementary visual information can be exploited when sufficient alignment is achieved.
However, further introducing ECG does not consistently yield additional gains and, in some tasks, leads to degraded performance.
We hypothesize that, for TRACE, this behavior reflects an important regime shift: although the method is motivated by conditional cross-modal estimation, expanding the modality set may primarily introduce a substantially higher degree of structured and modality-specific missingness, rather than merely increasing representational heterogeneity.
Under this regime, naïvely increasing the number of modalities does not necessarily guarantee monotonic performance improvements, as the fusion backbone must simultaneously accommodate more complex routing, alignment, and sparsity patterns.

\begin{figure}[t]
    \centering
    \includegraphics[width=\linewidth]{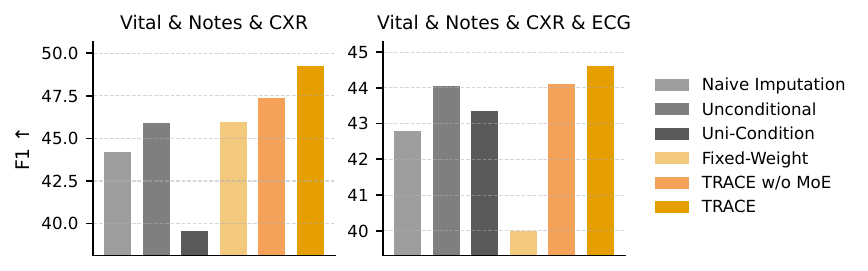}
    \caption{
    Ablation study of conditional diffusion and routing design on MIMIC-IV.
    Each panel corresponds to one modality setting.
    \textbf{Naive Imputation} follows FuseMoE-style naive imputation;
    \textbf{Unconditional} applies diffusion without conditioning on observed modalities;
    \textbf{Uni-Condition} conditions only on the observed part of the target modality;
    \textbf{Fixed-Weight} removes MoE routing by using fixed mixture weights;
    \textbf{TRACE w/o MoE} removes the MoE module entirely;
    \textbf{TRACE} denotes the full model.
    }
    \label{fig:ablation_f1}
\end{figure}

\subsection{Ablation Study}
\paragraph{Conditional Diffusion Block.}
To understand how TRACE realizes multimodal conditional estimation,
we conduct an ablation study that disentangles the roles of conditional diffusion
and adaptive expert routing.
We vary the conditioning strategy and routing mechanism,
with results summarized in Figure~\ref{fig:ablation_f1}.
Specifically, we compare naive deterministic imputation (Naive Imputation), unconditional diffusion (Unconditional),
target-only conditioning (Uni-Condition), and full cross-modal conditional diffusion (TRACE),
as well as variants that disable adaptive routing through fixed mixture weights (Fixed-Weight)
or by removing the MoE module entirely (TRACE w/o MoE).

Across all modality settings, naive imputation and unconditional diffusion underperform full conditional diffusion, showing that cross-modal conditioning is essential under multimodal missingness.
Uni-Condition improves only marginally over unconditional diffusion and remains below full TRACE, indicating that conditioning on complementary modalities is important.
Disabling adaptive routing through fixed weights or removing the MoE also hurts performance, so the best results come from combining conditional diffusion with data-dependent expert routing.

\begin{figure}[t]
    \centering
    \begin{subfigure}[t]{0.48\linewidth}
        \centering
        \includegraphics[width=\linewidth]{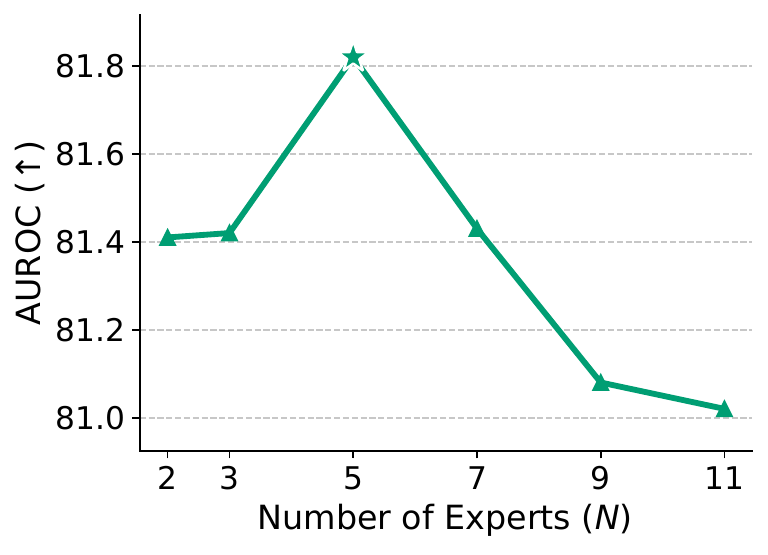}
        \caption{AUROC vs $N$ in MoE.}
        \label{fig:ablation_expert_rmse}
    \end{subfigure}
    \hfill
    \begin{subfigure}[t]{0.48\linewidth}
        \centering
        \includegraphics[width=\linewidth]{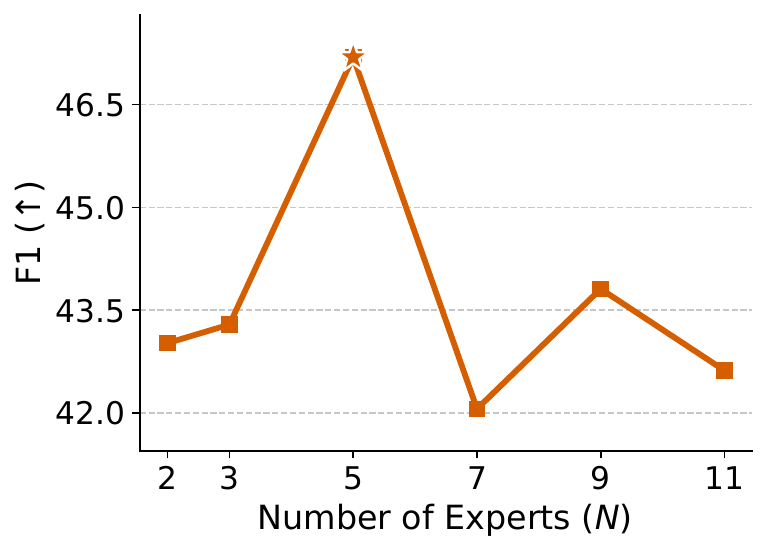}
        \caption{F1 vs $N$ in MoE.}
        \label{fig:ablation_expert_f1}
    \end{subfigure}
    \caption{
    Ablation on the number of experts $N$ in the conditional fusion MoE.
    Performance improves with increasing 
$N$ up to a moderate scale, with peak AUROC and F1 achieved at  $N = 5$.
    }
    \label{fig:ablation_expert}
\end{figure}

\paragraph{Number of Experts in Conditional Diffusion.}
We investigate the effect of the number of experts $N$ in the Mixture-of-Experts (MoE) component of the conditional diffusion module.
The expert cardinality controls the trade-off between denoising specialization and training stability: a larger $N$ increases routing capacity and allows experts to capture more diverse conditional patterns, while an overly large expert set may over-fragment training signals and lead to less reliable expert utilization.
We evaluate $N \in \{2,3,5,7,9,11\}$ and report validation AUROC and F1 score in Figure~\ref{fig:ablation_expert}.
As shown in the results, performance improves consistently as $N$ increases from 2 to 5 on both metrics, and degrades when further increasing $N$.
This non-monotonic trend suggests diminishing returns from additional experts beyond moderate cardinality,
as individual experts receive increasingly sparse supervision and become under-trained~\cite{shazeer2017outrageously}.
Both AUROC and F1 achieve their best performance at $N=5$, which we therefore adopt as the default configuration in all experiments.

\begin{table}[t]
    \centering
    \small
    \setlength{\tabcolsep}{5pt}
    \renewcommand{\arraystretch}{1.08}
    \caption{
    Efficiency analysis on the 48-IHM task of MIMIC-IV.
    Time is reported per sample at inference.
    }
    \label{tab:efficiency}
    \begin{tabular}{lccc}
    \toprule
    \textbf{Method} & \textbf{Time/sample} & \textbf{AUROC}$\uparrow$ & \textbf{F1}$\uparrow$ \\
    \midrule
    \textbf{FuseMoE} & 1.72 ms & 82.43 & 46.93 \\
    \textbf{TRACE (step=10)} & 228 ms & 83.05 & 47.59 \\
    \textbf{TRACE (step=50)} & 922 ms & \textbf{83.24} & \textbf{47.75} \\
    \bottomrule
    \end{tabular}
\end{table}

\paragraph{Efficiency Analysis.}
We further evaluate the computational cost of TRACE on the 48-IHM task, with results summarized in Table~\ref{tab:efficiency}.
Reducing the denoising horizon from 50 steps to 10 steps yields an approximately 4$\times$ speedup with only minor performance changes (AUROC 83.24 $\rightarrow$ 83.05, F1 47.75 $\rightarrow$ 47.59), while still outperforming FuseMoE on both metrics.
In addition, TRACE contains only 2.7M parameters (approximately 10.35MB of static memory), and its inference-time activation memory is about 170MB per sample, which remains tractable on standard GPU hardware.
For the clinical prediction settings considered here, where strict real-time response is typically not required, this overhead remains acceptable relative to the robustness gains under missingness.

\section{Discussion and Limitations}


\paragraph{Paradigm Generality vs.\ Domain Specialization.}
We note that on certain MIMIC-IV tasks, TRACE does not consistently outperform HAIM.
This observation is expected and reflects a fundamental difference in modeling scope rather than a limitation of the proposed paradigm.
In particular, for the 25-PHE task, HAIM achieves strong F1 performance, as this task emphasizes episode-level phenotype detection over the entire ICU stay and is less dependent on fine-grained temporal dynamics, while being sensitive to rare labels under the F1 metric.
HAIM is explicitly designed for this aggregation-dominated objective through full-stay feature aggregation and XGBoost-based classification, and accordingly performs strongly on tasks such as 25-PHE that favor global aggregation over temporal resolution.
Viewed from a different perspective, the performance gap highlights an inherent trade-off between domain-specialized architectures and paradigm-level generalization.
While task-specific pipelines such as HAIM can excel when their inductive biases closely match the objective, general frameworks like TRACE are designed to operate robustly across heterogeneous modalities, missingness patterns, and downstream tasks.
Bridging this gap through lightweight domain priors or task-aware conditioning within TRACE represents a promising direction for future work.


\paragraph{Scope of Missingness Modeling.}
While TRACE is formulated as a general paradigm for conditional representation estimation under multimodal missingness, the current instantiation focuses on settings where
modalities are partially observed rather than entirely absent.
Specifically, we target within-modality missingness, where each modality may be sparsely
or irregularly sampled but retains at least some observations.
For modality-level absence, our framework inherits the backbone capability of FuseMoE, whose routing mechanism can already adapt to unavailable modalities during fusion.
Accordingly, TRACE is intended to strengthen representation estimation when a modality is present but only partially observed.
This focus is also consistent with our empirical gains: TRACE is most effective when a modality remains partially observed and complementary modalities provide enough context to recover missing segments.
When an entire modality is absent, the problem shifts from conditional refinement to full substitution.
Extending conditional estimation from partial observation to explicit generation for entirely absent modalities remains an interesting direction for future work.

\paragraph{Rationale for Two-Stage Training.}
TRACE adopts a decoupled training strategy because the conditional diffusion stage is intended to learn task-agnostic representation estimation under partial observability before any downstream objective is imposed.
The two stages also optimize different targets: the diffusion module focuses on reconstructing plausible missing components under a generative loss, whereas the fusion stage focuses on discriminative downstream prediction, and directly coupling these objectives may bias estimation quality toward task-specific shortcuts.
In addition, incorporating iterative denoising into end-to-end optimization would substantially increase training cost and optimization instability, especially when the diffusion horizon is long.

\paragraph{Broader Impacts.}
This work may improve robustness in multimodal missing settings, which could be beneficial in applications like healthcare analysis.
Meanwhile, incorrect estimation may introduce downstream risks, so the method should be used with careful evaluation in high-stakes settings.

\section{Conclusions}
This work addresses the challenge of temporal misalignment and partial modality missingness
in multimodal time series foundation models.
We introduce \emph{TRACE}, a paradigm for cross-modal conditional
estimation under missingness and irregular sampling.
By modeling missing through conditional diffusion,
TRACE provides a principled alternative to naive imputation
and enables more consistent multimodal representations.
Empirical results across multiple multimodal time series settings demonstrate that
TRACE improves robustness and downstream performance,
highlighting the importance of explicit conditional estimation
as a core design principle for time series foundation.

\clearpage
\section*{Impact Statement}
This paper presents work whose goal is to advance the field of Machine Learning. There are many potential societal consequences of our work, none which we feel must be specifically highlighted here.

\bibliography{ref}
\bibliographystyle{icml2026}

\newpage
\appendix
\onecolumn
\section{Data, Tasks and Preprocessing}
\label{add:dataset}
Unless otherwise specified, our experimental protocol follows the experimental protocol introduced in FuseMoE \cite{han2024fusemoe}. This includes dataset configurations, preprocessing procedures, data splits, evaluation metrics, backbone architectures, and training hyperparameters. In the following, we briefly summarize the datasets and tasks used in our experiments.

\subsection{CMU-MOSI and CMU-MOSEI Datasets}
The CMU-MOSI dataset \cite{zadeh2016multimodal} is a multimodal sentiment analysis benchmark consisting of aligned text, audio, and visual modalities.
Specifically, it consists of 3,702 segments from 93 YouTube videos produced 
by 89 distinct speakers. 
These videos are further parsed into opinion and objective segments, 
and the accompanying sentiment with each of these segments (particularly subjective segments) is labeled. Beyond sentiment, 
each segment is also annotated for sentiment intensity and any gestures performed by the speaker. 
The CMU-MOSEI dataset \cite{zadeh2018multimodal} extends CMU-MOSI with a larger number of samples (23,453 segments) and more diverse speakers. 
Similar to CMU-MOSI, it contains aligned text, audio, and visual modalities.

Specifically, we perform two tasks on this dataset: 
\begin{itemize}
    \item \textbf{Sentiment intensity:} given an input segment and its associated additional modalities of data, determine the sentiment expressed in the segment on a scale of -3 (strongly negative) to 3 (strongly positive)
    \item \textbf{Sentiment polarity:} given an input segment together with its associated multimodal signals, predict the discrete sentiment label expressed in the segment. 
    Conceptually, sentiment polarity can be categorized into three classes: positive, negative, and neutral. 
    We adopt a binary polarity setting in our experiments, where neutral instances are merged into the negative class.

\end{itemize}

\subsection{MIMIC-IV Dataset}

The MIMIC-IV dataset \cite{johnson2020mimic} is a large-scale electronic health record (EHR) repository 
comprising patients who received critical care at the Beth Isreal Deaconess Medical Center. While patient inclusion
is determined by admissions to the emergency department or critical care units, captured data includes both 
ICU-specific information (as part of the \textit{icu} module), as well as comprehensive longitudinal health records 
(in the form of the \textit{hosp} module), enabling analyses that use long-term patient histories beyond 
a singular hospital visit.

In addition to the structured EHR data provided as part of its main module, MIMIC-IV includes several auxiliary modules
linked at the patient level, rendering the dataset inherently multi-modal. Specifically, patients from the core module 
can be associated with imaging data consisting of chest x-rays (MIMIC-CXR-JPG \cite{johnson2019mimic}), text in the form of clinical narratives 
(MIMIC-IV-Note\cite{PhysioNet-mimic-iv-note-2.2}) and physiological time series data in the form of electrocardiograms (MIMIC-IV-ECG\cite{gow2023mimic}). 

Our current experiments primarily focus on leveraging the inherent multimodality of this full data with additional 
modules to support three common EHR-based tasks, matching those used as part of FuseMoE \cite{han2024fusemoe}:
\begin{itemize}
    \item \textbf{48 hour in-hospital mortality prediction:} The model is tasked to make a binary prediction of whether 
    a patient will die in-hospital given the initial 48 hours of data post-admission. 
    \item \textbf{25-type phenotype classification:} The model is provided with all the data prior to discharge, and asked
    to predict whether or not one of 25 broad critical care conditions (based on the Elixhauser comorbidity measures \cite{elixhauser1998comorbidity}) will be present upon discharge. 
    \item \textbf{Length of stay prediction:} The model is given all data accumulated within the first 48 hours
    post-admission and is tasked to predict whether the patient will be discharged within the following 48 hours.
\end{itemize}

\subsection{Motivational Dataset}
\label{sec:toy_example}
We design a synthetic multimodal time series dataset that enables controlled analysis of
temporal dynamics, frequency decomposition, and conditional information abstraction.
We define a forecasting task on the dataset, in which models are required to infer future temporal
dynamics from partially observed histories, using the fully observed sequence as oracle
supervision.
Each sample is first generated as a fully observed multivariate time series together with
two lossy conditional modalities summarizing complementary frequency regimes of the
underlying generative process.
This fully observed construction allows us to isolate conditional generation and inference
under explicitly non-invertible conditioning.

Starting from the complete data, we subsequently introduce controlled missingness as a
separate evaluation mechanism using structured block masking over consecutive time steps and
feature dimensions, with varying block sizes to obtain multiple missing ratios.

Each sample is composed of \textbf{three modalities}:
\begin{itemize}
\item A multivariate time series $\mathbf{X} \in \mathbb{R}^{L \times D}$, where $L=50$ and $D=30$.

\item A low-frequency + trend condition embedding $\mathbf{c}_{\text{low}} \in \mathbb{R}^{16}$.

\item A mid/high-frequency + burst condition embedding $\mathbf{c}_{\text{high}} \in \mathbb{R}^{16}$.

\end{itemize}

The dataset contains 20{,}000 training samples, 2{,}000 validation samples, and
2{,}000 test samples.
\paragraph{Time Series Modality.}
The synthetic time series exhibits multi-resolution structure, non-stationarity,
and localized burst patterns, resembling clinical time series data such as
MIMIC-style signals.
All feature dimensions are statistically correlated yet not directly disentangled.

\vspace{0.5em}
The $30$ latent feature dimensions are implicitly divided into
$8$ low-frequency + trend dimensions,
$6$ mid-frequency dimensions,
and $16$ high-frequency + burst dimensions.
After generation, all dimensions are mixed in the observation space.

\begin{itemize}
    \item Low-frequency + trend components.
    These components control global temporal trends and smooth variations.
    For dimension $d$, the signal is defined as
    \begin{equation}
        x_d^{\text{low}}(t)
        =
        a_d^{\text{low}} \sin\!\left(2\pi f_d^{\text{low}} t + \phi_d^{\text{low}}\right)
        + s_d (t - 0.5)
        + q_d (t - 0.5)^2,
    \end{equation}
    where the frequency $f_d^{\text{low}} \sim \mathcal{U}(0.6, 2.0)$,
    amplitude $a_d^{\text{low}} \sim \mathcal{U}(0.6, 1.5)$,
    linear trend $s_d \sim \mathcal{U}(-1.0, 1.0)$,
    quadratic trend $q_d \sim \mathcal{U}(-0.3, 0.3)$,
    and phase $\phi_d^{\text{low}} \sim \mathcal{U}(0, 2\pi)$.

    \item Mid-frequency components.
    These components provide medium-scale temporal oscillations.
    For dimension $d$, we define
    \begin{equation}
        x_d^{\text{mid}}(t)
        =
        a_d^{\text{mid}} \sin\!\left(2\pi f_d^{\text{mid}} t + \phi_d^{\text{mid}}\right),
    \end{equation}
    where the frequency $f_d^{\text{mid}} \sim \mathcal{U}(2.5, 5.0)$,
    the amplitude $a_d^{\text{mid}} \sim \mathcal{U}(0.4, 1.2)$,
    and the phase $\phi_d^{\text{mid}} \sim \mathcal{U}(0, 2\pi)$
    are sampled independently across dimensions.
    
    \item High-frequency + burst components.
    High-frequency components capture rapid oscillations,
    \begin{equation}
        \tilde{x}_d^{\text{high}}(t)
        =
        a_d^{\text{high}} \sin\!\left(2\pi f_d^{\text{high}} t + \phi_d^{\text{high}}\right)
        +
        0.3\, a_d^{\text{high}}
        \cos\!\left(2\pi (1.7 f_d^{\text{high}}) t + \frac{\phi_d^{\text{high}}}{2}\right),
    \end{equation}
    where the frequency $f_d^{\text{high}} \sim \mathcal{U}(6.0, 11.0)$,
    the amplitude $a_d^{\text{high}} \sim \mathcal{U}(0.2, 0.9)$,
    and the phase $\phi_d^{\text{high}} \sim \mathcal{U}(0, 2\pi)$.
    With probability $0.7$, a burst event is injected using a Gaussian bump:
    \begin{equation}
        g(t)
        =
        \exp\!\left(
        - \frac{(t - \mu)^2}{2\sigma^2}
        \right),
    \end{equation}
    where the burst location $\mu \sim \mathcal{U}(0.15, 0.85)$,
    width $\sigma \sim \mathcal{U}(0.05, 0.1)$,
    and amplitude $A \sim \mathcal{U}(0.8, 2.2)$.
    The burst contribution is given by
    \begin{equation}
        x_d^{\text{burst}}(t)
        =
        A g(t),
    \end{equation}
    and the final high-frequency signal is
    $x_d^{\text{high}}(t) = \tilde{x}_d^{\text{high}}(t) + x_d^{\text{burst}}(t)$.
    
    \item Frequency-wise aggregation.
    After generating the low-frequency, mid-frequency, and high-frequency (with burst)
    components, we aggregate them to form the complete latent multivariate signal.
    Specifically, for each feature dimension $d$, the latent signal is defined as
    \begin{equation}
    x_d(t)
    =
    \begin{cases}
    x_d^{\text{low}}(t), & d \in \mathcal{D}_{\text{low}}, \\
    x_d^{\text{mid}}(t), & d \in \mathcal{D}_{\text{mid}}, \\
    x_d^{\text{high}}(t), & d \in \mathcal{D}_{\text{high}},
    \end{cases}
    \end{equation}
    where $\mathcal{D}_{\text{low}}$, $\mathcal{D}_{\text{mid}}$, and $\mathcal{D}_{\text{high}}$
    denote the index sets corresponding to the low-, mid-, and high-frequency feature
    groups, respectively.

    \item Cross-channel correlation via shared drift.
    To introduce statistical dependency across feature dimensions,
    we add a shared temporal drift:
    \begin{equation}
    d(t)
    =
    a^{\text{drift}}
    \sin\!\left(2\pi f^{\text{drift}} t + \phi^{\text{drift}}\right),
    \end{equation}
    where
    $a^{\text{drift}} \sim \mathcal{U}(0.05, 0.35)$,
    $f^{\text{drift}} \sim \mathcal{U}(0.1, 0.5)$,
    and $\phi^{\text{drift}} \sim \mathcal{U}(0, 2\pi)$.
    After frequency-wise aggregation, the latent signal is lifted to vector form
    $\mathbf{x}(t) = (x_1(t), \dots, x_D(t))^\top$,
    and the drift is added uniformly across all dimensions as
    \begin{equation}
    \tilde{\mathbf{x}}(t)
    =
    \mathbf{x}(t)
    +
    d(t)\,\mathbf{1},
    \end{equation}
    where $\mathbf{1} \in \mathbb{R}^D$ denotes the all-ones vector.
    
    \item Cross-feature mixing.
    A random mixing matrix
    $\mathbf{M} \in \mathbb{R}^{D \times D}$
    is then applied at the vector level as
    \begin{equation}
    \mathbf{x}'(t)
    =
    \mathbf{M}\,\tilde{\mathbf{x}}(t),
    \end{equation}
    ensuring that each observed dimension is a mixture of multiple latent
    frequency components and preventing trivial feature-wise disentanglement.
    
    \item Observation noise.
    Finally, stacking $\mathbf{x}'(t)$ over all time steps yields
    $\mathbf{X} = [\mathbf{x}'(1), \dots, \mathbf{x}'(L)]^\top \in \mathbb{R}^{L \times D}$,
    and additive Gaussian noise is applied as
    \begin{equation}
    \mathbf{X}
    \leftarrow
    \mathbf{X}
    +
    \boldsymbol{\epsilon},
    \qquad
    \boldsymbol{\epsilon} \sim \mathcal{N}(0, 0.03^2).
    \end{equation}

\end{itemize}
\paragraph{Conditional Modalities.}
We construct two complementary conditional modalities,
$\mathbf{c}_{\text{low}}$ and $\mathbf{c}_{\text{high}}$,
to summarize different frequency regimes of the generative process.
Importantly, both conditions are derived from the latent generation parameters
rather than the observed time series, and are intentionally designed to be lossy and
non-invertible.

The low-frequency condition $\mathbf{c}_{\text{low}} \in \mathbb{R}^{16}$ statistically
compresses the parameters of low-frequency sinusoidal components and their associated
trend terms.
It aggregates summary statistics of amplitudes, frequencies, phases, and linear/quadratic
trend coefficients, including means, standard deviations, and cross-statistics, and
additionally encodes the shared drift parameters to capture global cross-channel trends.
The mid/high-frequency condition $\mathbf{c}_{\text{high}} \in \mathbb{R}^{16}$ summarizes
faster oscillatory behavior and localized burst events by aggregating statistics of
mid- and high-frequency sinusoidal parameters, explicitly encoding burst metadata(e.g., presence, temporal location, width, and amplitude)
, and incorporating band-energy descriptors
to reflect multi-scale activity.

For both conditions, the aggregated statistics are first concatenated into
condition-specific summary vectors
$\mathbf{u}_1, \mathbf{u}_2 \in \mathbb{R}^{28}$.
To prevent an invertible mapping between the condition and the underlying signal,
the summary vector is then compressed via a fixed random projection with additive noise:
\begin{equation}
\label{eq:cond_lowtrend}
\mathbf{c}_{\text{low}} = \mathbf{P}_1 \mathbf{u}_1 + \boldsymbol{\epsilon}_1,
\qquad
\boldsymbol{\epsilon}_1 \sim \mathcal{N}(0, 0.02^2 \mathbf{I}),
\end{equation}
\begin{equation}
\label{eq:cond_midhighburst}
\mathbf{c}_{\text{high}} = \mathbf{P}_2 \mathbf{u}_2 + \boldsymbol{\epsilon}_2,
\qquad
\boldsymbol{\epsilon}_2 \sim \mathcal{N}(0, 0.02^2 \mathbf{I}),
\end{equation}
where $\mathbf{P}_1, \mathbf{P}_2 \in \mathbb{R}^{16 \times 28}$ are fixed random projection
matrices shared across the dataset.
In both cases, additional nuisance dimensions with small Gaussian noise are
injected into $\mathbf{u}_1$ and $\mathbf{u}_2$ prior to projection to further destroy
invertibility.
As a result, the conditional embeddings provide strongly correlated global priors over
temporal structure and burst activity, while discarding local information required to
reconstruct the full time series.

\paragraph{Controlled Missingness via Block Masking.}
Starting from the fully observed data, we introduce controlled missingness
as a separate evaluation layer using a structured block-masking scheme.
Specifically, we mask contiguous blocks spanning consecutive time steps
and feature dimensions, thereby inducing temporally and cross-dimensionally
correlated missing patterns.
By varying the size of the masked blocks, we obtain multiple missing ratios,
which allows us to systematically assess model robustness under increasing
levels of data sparsity.
Such structured missingness closely reflects common failure modes in
real-world time series data, where sensor outages, acquisition delays,
or modality dropouts often lead to temporally contiguous and correlated
missing segments.

\paragraph{Signal-Level Comparison under Missingness.}
We first compare TRACE and FuseMoE at the signal level under the controlled missingness setting,
which directly evaluates imputation accuracy.
Starting from fully observed time series, we apply structured block masking to induce missing values
and assess the quality of the imputed signals produced by each method.
The comparison focuses exclusively on missing entries, thereby isolating imputation performance
from trivially observed values.

Specifically, we measure the discrepancy between the imputed values and the ground-truth signals
at missing positions using the mean absolute error at missing entries (MAE@Missing).
Formally, for a sample $i$, let
$\Omega_i = \{(t,d) \mid m_i(t,d)=0\}$
denote the set of missing time--dimension indices, where $m_i(t,d)$ is the corresponding observation mask.
The MAE@Missing is defined as
\begin{equation}
\mathrm{MAE@Missing}
= \frac{1}{|\Omega_i|}
\sum_{(t,d)\in\Omega_i}
\left| \hat{x}_i(t,d) - x_i(t,d) \right|,
\end{equation}
where $x_i(t,d)$ and $\hat{x}_i(t,d)$ denote the ground-truth and imputed signal values, respectively.
This metric provides a direct and focused evaluation of how effectively each method reconstructs
unobserved signals under structured missingness.

\paragraph{Representation-Level Comparison under Missingness.}
While signal-level metrics quantify imputation accuracy at missing entries,
the advantage of diffusion-based imputation over naive imputation is expected
and, to some extent, trivial.
A more subtle question is to what extent the artifacts introduced by naive imputation
can be fully absorbed or corrected by the subsequent fusion layers,
thereby yielding comparable internal temporal representations.

To address this question, we further conduct a representation-level comparison
under the same controlled missingness setting.
Such a comparison is non-trivial, as representations learned from differently
imputed inputs do not naturally reside in a shared feature space:
models trained on different imputations may exhibit representation collapse,
scale mismatch, or arbitrary rotations induced by optimization and
hyperparameter choices, making direct comparison ill-posed.

Specifically, we first obtain oracle representations by applying the same
fusion architecture to fully observed inputs, which defines a common
reference representation space.
The fusion model is trained once using fully observed multimodal time series
data following the TRACE architecture, with no missing values.
On the test set, we first apply this trained fusion model to the fully observed
inputs to obtain oracle sequence-level representations, which serve as the
ground-truth reference in the latent space.
Keeping the fusion layer parameters fixed, we then feed the imputed test inputs
produced by TRACE and the FuseMoE-based naive imputation into this fusion layer
and extract the resulting sequence-level embeddings after the multi-time
attention module (mTAND).
The imputed representations are subsequently compared against the
corresponding oracle representations obtained from fully observed inputs.

By operating at the representation level, this evaluation decouples imputation quality
from downstream prediction heads and focuses solely on the fidelity of learned temporal representations.

The comparison is conducted in the latent space using cosine similarity, which measures
the angular alignment between representations while being invariant to their scale.
Formally, given two sequence-level representations $\mathbf{h}$ and $\mathbf{h}^{\ast}$,
the cosine similarity is defined as
\begin{equation}
\label{eq:cosine_similarity}
s_{\mathrm{cos}}(\mathbf{h}, \mathbf{h}^{\ast})
=
\frac{\langle \mathbf{h}, \mathbf{h}^{\ast} \rangle}
{\|\mathbf{h}\|_2 \, \|\mathbf{h}^{\ast}\|_2},
\end{equation}
where $\langle \cdot, \cdot \rangle$ denotes the inner product.
A higher cosine similarity indicates better alignment between the inferred representation
and the oracle representation derived from complete data, independent of their magnitude.

Overall, this construction yields a controlled benchmark in which
the observed time series and conditional embeddings are generated in a fully specified manner,
and missingness is introduced in a structured and reproducible way,
enabling systematic evaluation of conditional generation and imputation methods
under non-invertible conditioning.

\section{Implementation Details}
\subsection{Observation Mask}
\label{app:masking}
We consider multimodal time series data under partial observability.
For each modality $m$, we denote the preprocessed representation as
$x^{(m)} \in \mathbb{R}^{L \times d_m}$, where $L$ is the temporal length
and $d_m$ is the modality-specific feature dimension.
Missingness is modeled via a binary observation mask
$r^{(m)} \in \{0,1\}^{L \times d_m}$,
where $r^{(m)}_{ij}=1$ indicates an observed entry and $r^{(m)}_{ij}=0$ indicates a missing entry.

Using the observation mask, the observed and unobserved components of modality $m$
are given by
\begin{equation}
x^{(m)}_{\mathcal{O}} = r^{(m)} \odot x^{(m)},
\qquad
x^{(m)}_{\mathcal{U}} = (1-r^{(m)}) \odot x^{(m)}
\end{equation}

where $\odot$ denotes element-wise multiplication.
Our goal is to estimate the unobserved components $x^{(m)}_{\mathcal{U}}$
conditioned on all available information.

The observation mask is applied consistently across modalities.
For auxiliary modalities, the corresponding observation masks are applied when constructing
conditioning signals, such that only observed entries contribute to the multimodal conditioning context.

For a target modality $m$, the diffusion-based estimator is tasked with generating values for
the unobserved components $x^{(m)}_{\mathcal{U}}$ conditioned on the observed entries
$x^{(m)}_{\mathcal{O}}$ and multimodal context.
Accordingly, the diffusion latents $\{z^{(m)}_t\}_{t=0}^T$ parameterize the stochastic evolution of the
missing components, while observed entries are treated as fixed conditions.

\subsection{Denoising Diffusion Probabilistic Models}

We briefly review denoising diffusion probabilistic models (DDPM)~\cite{ho2020denoising},
which serve as the foundation of our conditional diffusion formulation.
DDPM defines a latent-variable generative model
based on a forward noising process and a learned reverse denoising process.

Let $z^{(m)}_0 \in \mathbb{R}^{L \times d_m}$ denote a clean representation
in the sample space of modality $m$.
The forward diffusion process gradually perturbs $z^{(m)}_0$ with Gaussian noise
through a Markov chain defined as
\begin{equation}
\label{eq:ddpm_forward}
q\!\left(z^{(m)}_{1:T} \mid z^{(m)}_0\right)
:=
\prod_{t=1}^{T}
q\!\left(z^{(m)}_t \mid z^{(m)}_{t-1}\right),
\end{equation}
where
$q\!\left(z^{(m)}_t \mid z^{(m)}_{t-1}\right)
=
\mathcal{N}\!\left(
z^{(m)}_t;
\sqrt{1-\beta_t}\, z^{(m)}_{t-1},
\beta_t I
\right)$
,
and $\{\beta_t\}_{t=1}^{T}$ is a predefined variance schedule.
Equivalently, the diffusion trajectory admits the closed-form reparameterization
\begin{equation}
\label{eq:ddpm_reparam}
z^{(m)}_t
=
\sqrt{\bar{\alpha}_t}\, z^{(m)}_0
+
\sqrt{1-\bar{\alpha}_t}\, \epsilon,
\qquad
\epsilon \sim \mathcal{N}(0,I),
\end{equation}
with $\bar{\alpha}_t = \prod_{i=1}^{t}(1-\beta_i)$.

The reverse generative process aims to recover $z^{(m)}_0$ from Gaussian noise
and is defined as
\begin{equation}
\label{eq:ddpm_reverse}
p_\theta\!\left(z^{(m)}_{0:T}\right)
:=
p\!\left(z^{(m)}_T\right)
\prod_{t=1}^{T}
p_\theta\!\left(z^{(m)}_{t-1} \mid z^{(m)}_t\right),
\qquad
z^{(m)}_T \sim \mathcal{N}(0,I),
\end{equation}
where each reverse transition is parameterized as
\begin{equation}
p_\theta\!\left(z^{(m)}_{t-1} \mid z^{(m)}_t\right)
=
\mathcal{N}\!\left(
z^{(m)}_{t-1};
\mu_\theta\!\left(z^{(m)}_t,t\right),
\sigma_t^2 I
\right).
\end{equation}

Following the noise-prediction parameterization in DDPM,
the mean $\mu_\theta$ is expressed via a denoising network
$\epsilon_\theta$ as
\begin{equation}
\mu_\theta\!\left(z^{(m)}_t,t\right)
=
\frac{1}{\sqrt{\alpha_t}}
\left(
z^{(m)}_t
-
\frac{1-\alpha_t}{\sqrt{1-\bar{\alpha}_t}}
\epsilon_\theta\!\left(z^{(m)}_t,t\right)
\right),
\end{equation}
where $\alpha_t = 1-\beta_t$.
The variance $\sigma_t^2$ is determined by the predefined noise schedule,
following standard DDPM formulations.

Under this parameterization, the denoising network $\epsilon_\theta$
is trained to predict the Gaussian noise added in the forward process
by minimizing the objective
\begin{equation}
\label{eq:ddpm_objective}
\mathcal{L}_{\mathrm{DDPM}}(\theta)
=
\mathbb{E}_{z^{(m)}_0,\epsilon,t}
\left[
\left\|
\epsilon
-
\epsilon_\theta\!\left(z^{(m)}_t,t\right)
\right\|_2^2
\right].
\end{equation}
where $\epsilon \sim \mathcal{N}(0,I)$,
$t$ is sampled uniformly from $\{1,\dots,T\}$,
and $z^{(m)}_t$ is obtained from $z^{(m)}_0$ via the forward diffusion process
in Eq.~\eqref{eq:ddpm_reparam}.

\subsection{Conditional Diffusion for Imputation}
We consider the imputation problem under partial observability,
where the goal is to estimate missing components of a target modality
conditioned on available information.
Specifically, for a target modality $m$, we aim to model the conditional distribution
$p_\theta\!\left(x^{(m)}_{\mathcal{U}} \mid x^{(m)}_{\mathcal{O}}, c^{(m)}\right)$,
where $x^{(m)}_{\mathcal{O}}$ and $x^{(m)}_{\mathcal{U}}$ denote the observed and missing
components of the target modality, and $c^{(m)}$ represents the aggregated multimodal
conditioning context constructed from auxiliary modalities.

To model this conditional distribution, we define a conditional diffusion process
over the target-modality latent variables.
Following DDPM, the conditional reverse process is given by
\begin{equation}
p_\theta\!\left(z^{(m)}_{0:T} \mid c^{(m)}\right)
=
p\!\left(z^{(m)}_T\right)
\prod_{t=1}^{T}
p_\theta\!\left(z^{(m)}_{t-1} \mid z^{(m)}_t, c^{(m)}\right),
\end{equation}
where $z^{(m)}_T \sim \mathcal{N}(0,I)$.
Each reverse transition is parameterized as
\begin{equation}
p_\theta\!\left(z^{(m)}_{t-1} \mid z^{(m)}_t, c^{(m)}\right)
=
\mathcal{N}\!\left(
z^{(m)}_{t-1};
\mu_\theta\!\left(z^{(m)}_t,t \mid c^{(m)}\right),
\sigma_t^2 I
\right).
\end{equation}

Under the noise-prediction parameterization, the conditional denoising network
$\epsilon_\theta(\cdot \mid c^{(m)})$ is trained by minimizing
\begin{equation}
\mathcal{L}_{\mathrm{diff}}(\theta)
=
\mathbb{E}_{z^{(m)}_0,\epsilon,t}
\left[
\left\|
\epsilon
-
\epsilon_\theta\!\left(z^{(m)}_t,t \mid c^{(m)}\right)
\right\|_2^2
\right],
\end{equation}
where $z^{(m)}_t$ is obtained via the forward diffusion process.

\section{Baseline Methods}

\subsection{MISTS}
\citet{zhang2023improving} introduced a framework designed to handle irregularity in multivariate time series and clinical notes. MISTS leverages Time2Vec-based embeddings \citep{Kazemi2019Time2VecLA} to encode irregular temporal information and uses self-attention and cross-attention layers to fuse modalities. 
\subsection{MulT}
MulT, proposed by \citet{tsai2019multimodal}, is a transformer-based multi-modal model that fuses unaligned sequences via directional crossmodal attention instead of temporal alignment. By stacking pairwise crossmodal transformers and applying self-attention for temporal aggregation, MulT captures long-range cross modal dependencies and enables effective multimodal fusion without requiring predefined alignment. 
\subsection{MAG}
MAG  \cite{rahman2020integrating} is an attachment-based framework to incorporate nonverbal modalities into pretrained language models such as BERT and XLnet during fine tuning. MAG keeps the original architecture of the pretrained language model and introduces minimal additional parameters, enabling efficient multimodal adaptation without modifying the transformer backbone. 
\subsection{TFN}
TFN consists of three main components. It first encodes each modalities (language, visual, and acoustic) using Modality Embedding Subnetworks and then applies a Tensor Fusion Layer to jointly represent all combinations of cross-modal interactions. A Sentiment Inference Subnetwork is implemented to perform the final prediction \citep{zadeh2017tensor} 
\subsection{HAIM}
HAIM is a multimodal framework for healthcare prediction that integrates heterogeneous patient data from multiple data sources. Each data modality, including tabular data, time series data, clinical notes, medical images, and other resources, is processed independently through modality-specific embedding pipelines. These generated embeddings are then concatenated into a unified fusion representation, which is used for downstream tasks \citep{soenksen2022integrated}. 
\subsection{FuseMoE}
FuseMoE is a MoE based framework designed for multimodal fusion of  heterogeneous multimodal data with varying modalities. This method uses sparsely gated MoE layers in the fusion stage to route each modality to specialized experts. This design enables FuseMoE to not be limited to a fixed set of modalities, while handling missing modalities by dynamically adjusting expert contributions based on input \citep{han2024fusemoe}. 

\begin{table*}[thbp]
\centering
\small
\setlength{\tabcolsep}{3.5pt}
\renewcommand{\arraystretch}{1.05}
\caption{Hyperparameter configurations used for TRACE across different datasets and tasks.}
\begin{tabular}{c|c|ccccc}
\toprule
\multirow{2}{*}{\textbf{Hyperparameter Type}} & \multirow{2}{*}{\textbf{Parameter Name}} & \multicolumn{3}{c}{\textbf{Dataset}} \\
& & MIMIC-IV & CMU-MOSI & CMU-MOSEI \\
\midrule
\multirow{10}{*}{Conditional Diffusion} 
& Batch size & 16 & 16 & 16 \\
& Number of experts & 5 & 5 & 5 \\
& N samples & 20 & 1 & 1 \\
& Diffusion time steps & 50 & 50 & 50 \\
& Number of denoising layers & 4 & 4 & 4 \\
& Mask ratio & 20\% & 20\% & 20\% \\
& Diffusion embedding dimension & 128 & 128 & 128 \\
& Feature embedding dimension & 64 & 64 & 64 \\
& Time embedding dimension & 128 & 128 & 128 \\
& Multimodal context dimension & 128 & 128 & 128 \\

\midrule
\multirow{7}{*}{MoE Fusion Layer} 
& Training epochs & 32 & 40 & 20 \\
& Number of MoE layers & 3 & 3 & 1 \\
& Hidden size & 512 & 512 & 512 \\
& Number of experts & 16 & 8 & 16 \\
& Top k & 4 & 4 & 4 \\
& Router Type & Joint & Per-mod & Joint \\
& Gating Function & Laplace & Laplace & Laplace \\

\midrule
\multirow{6}{*}{Other Parameters} 
& Random seed & [42, 52, 72] & [42, 52, 72] & [42, 52, 72] \\
& BERT or T5 learning rate & 2e-5 & 3e-4 & 3e-4 \\
& Default learning rate & 4e-4 & 1e-4 & 1e-4 \\
& Number of attention heads & 8 & 8 & 8 \\
& Processed sequence length & 48 & 50 & 50 \\
& Attention embedding dimension & 128 & 32 & 32 \\
\bottomrule
\end{tabular}

\label{tab:hyperparameters}
\end{table*}

\section{Implementation Details}
\subsection{Computational Resources}
All experiments were performed using 8 NVIDIA RTX 6000 Ada GPUs, each with 48 GB of memory. All training runs utilized a single GPU.

\subsection{Hyper-Parameters}
The complete set of hyperparameters employed in our experiments is provided in Table \ref{tab:hyperparameters}. These include detailed settings for each part of TRACE, covering Multimodal Conditional Diffusion, MoE Fusion Layer, and other related parameters.

\subsection{Dataset-Specific Instantiation of Conditional Estimation}
While TRACE is formulated as a general paradigm for conditional estimation across modalities, the specific instantiation varies across datasets in our experiments.
In the MIMIC-IV dataset, we apply conditional estimation to the time-series modality only, following the experimental protocol of FuseMoE to enable fair comparison.
In contrast, for the CMU-MOSI and CMU-MOSEI datasets, where all modalities are continuously observed and aligned at the segment level, we extend conditional estimation to multiple modalities.

\begin{table*}[thbp]
\centering
\small
\setlength{\tabcolsep}{3.5pt}
\renewcommand{\arraystretch}{1.05}
\caption{Ablation study on N-Sample during diffusion imputation on the 48-IHM task in the MIMIC-IV dataset.}
\begin{tabular}{l|cc|cc|cc}
\toprule
\textbf{N-Sample}
& \multicolumn{2}{c|}{\textbf{TS \& Text}}
& \multicolumn{2}{c|}{\textbf{TS \& CXR \& Text}}
& \multicolumn{2}{c}{\textbf{TS \& CXR \& Text \& ECG}} \\
\cmidrule(lr){2-3}\cmidrule(lr){4-5}\cmidrule(lr){6-7}
& AUROC$\uparrow$ & F1$\uparrow$
& AUROC$\uparrow$ & F1$\uparrow$
& AUROC$\uparrow$ & F1$\uparrow$ \\
\midrule

1
& 80.09 & 47.11
& 83.02 & 45.49
& 80.68 & 43.76 \\

5
& 81.95 & 46.60
& 83.01 & 47.79
& 78.77 & 40.12 \\

10
& 83.31 & 47.21
& 83.86 & 45.82
& 81.88 & 43.15 \\

20
& \textbf{83.43} & 46.93
& 83.84 & \textbf{49.22}
& \textbf{82.02} & \textbf{44.60} \\

30
& 82.80 & \textbf{47.71}
& \textbf{83.94} & 47.27
& 81.07 & 41.72 \\

\bottomrule
\end{tabular}

\label{tab:ab2}
\end{table*}
\section{Additional Results}
\subsection{Ablation Study of N-Sample During Diffusion Imputation}
Table \ref{tab:ab2}. reports an ablation study on the number of samples used during conditional diffusion
imputation on the 48-hour in-hospital mortality (48-IHM) task.
Across different modality combinations, increasing the number of diffusion samples generally
improves AUROC, indicating that aggregating multiple stochastic imputations leads to more stable
and reliable estimates of missing representations.
However, the gains are not strictly monotonic, and performance saturates or slightly degrades
when the sample size becomes large.

In the two-modality setting (TS \& Text), performance improves steadily as the number of samples
increases, with the best AUROC achieved at moderate sampling budgets.
In the three-modality setting (TS \& CXR \& Text), larger sample sizes further benefit performance,
suggesting that additional stochastic estimates help mitigate increased modality heterogeneity.
By contrast, in the four-modality setting (TS \& CXR \& Text \& ECG), the optimal performance
is attained with fewer samples, reflecting the higher variance and complexity introduced by
additional modalities.

Overall, these results suggest that while multi-sample diffusion imputation can enhance robustness,
the optimal sampling budget depends on the modality composition and the degree of multimodal
heterogeneity, and moderate sampling is often sufficient in practice.

\begin{figure*}[t]
    \centering
    \setlength{\tabcolsep}{2pt}
    \begin{tabular}{cccc}
        \includegraphics[width=0.32\linewidth]{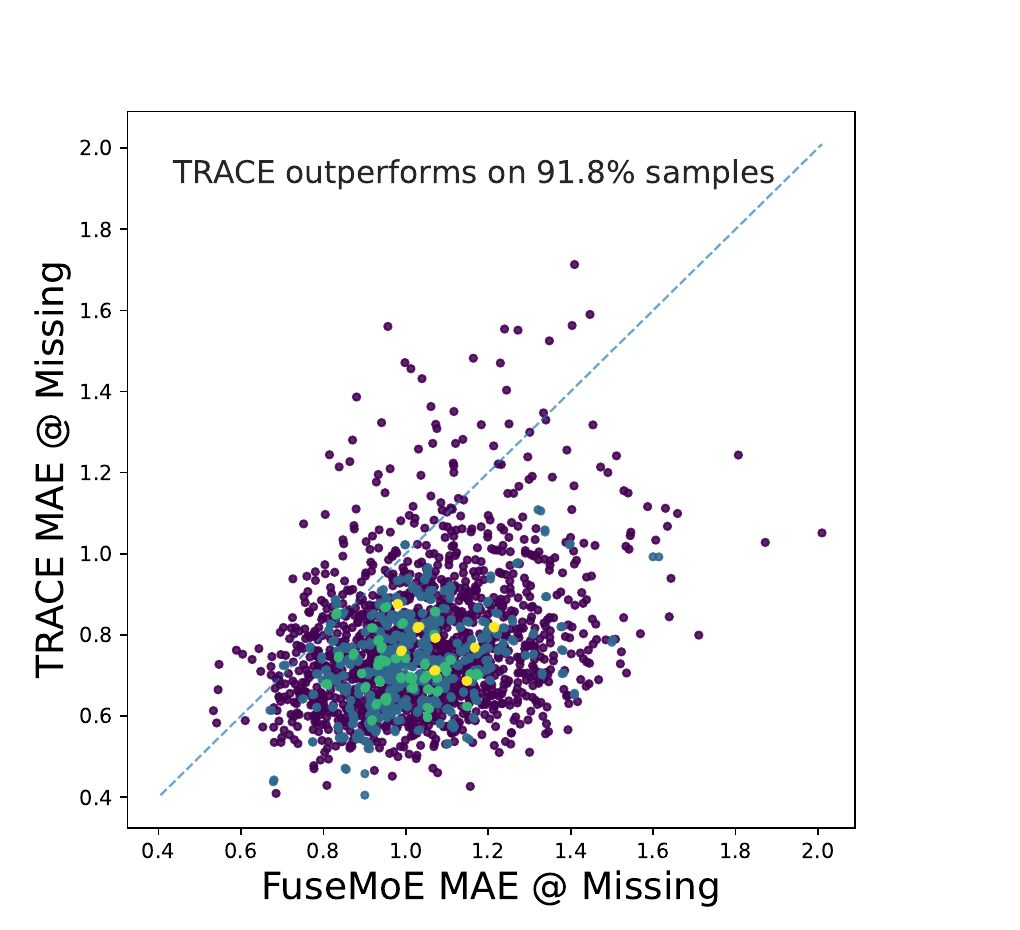}  &
        \includegraphics[width=0.32\linewidth]{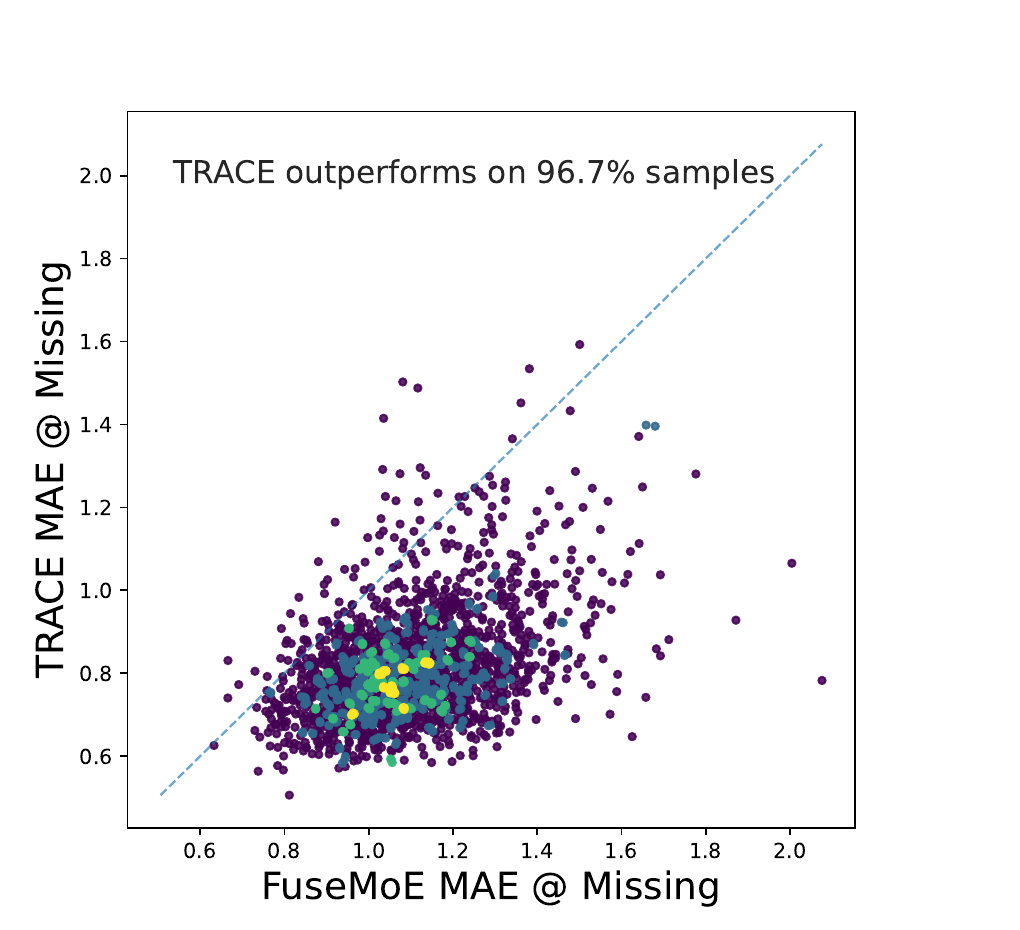}  &
        \includegraphics[width=0.32\linewidth]{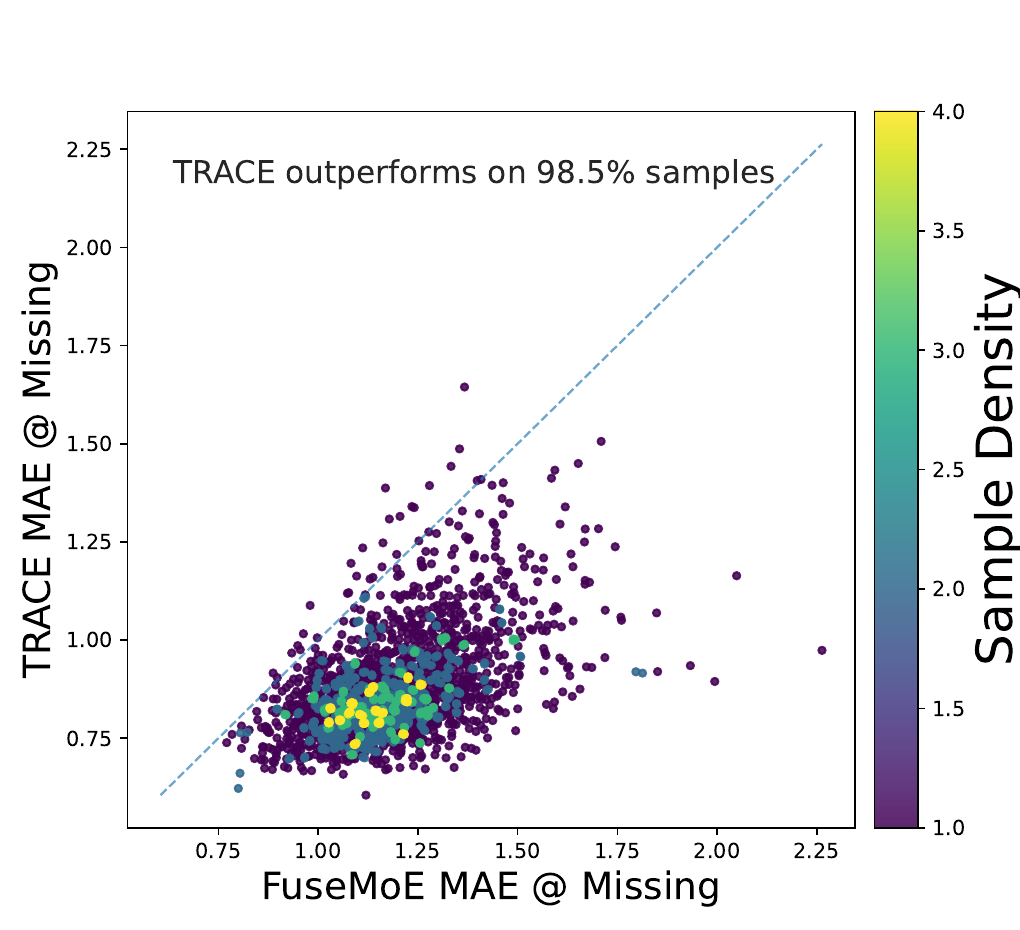} &\\
        \small MR=5\% & \small MR=10\% & \small MR=20\% \\[4pt]

        \includegraphics[width=0.32\linewidth]{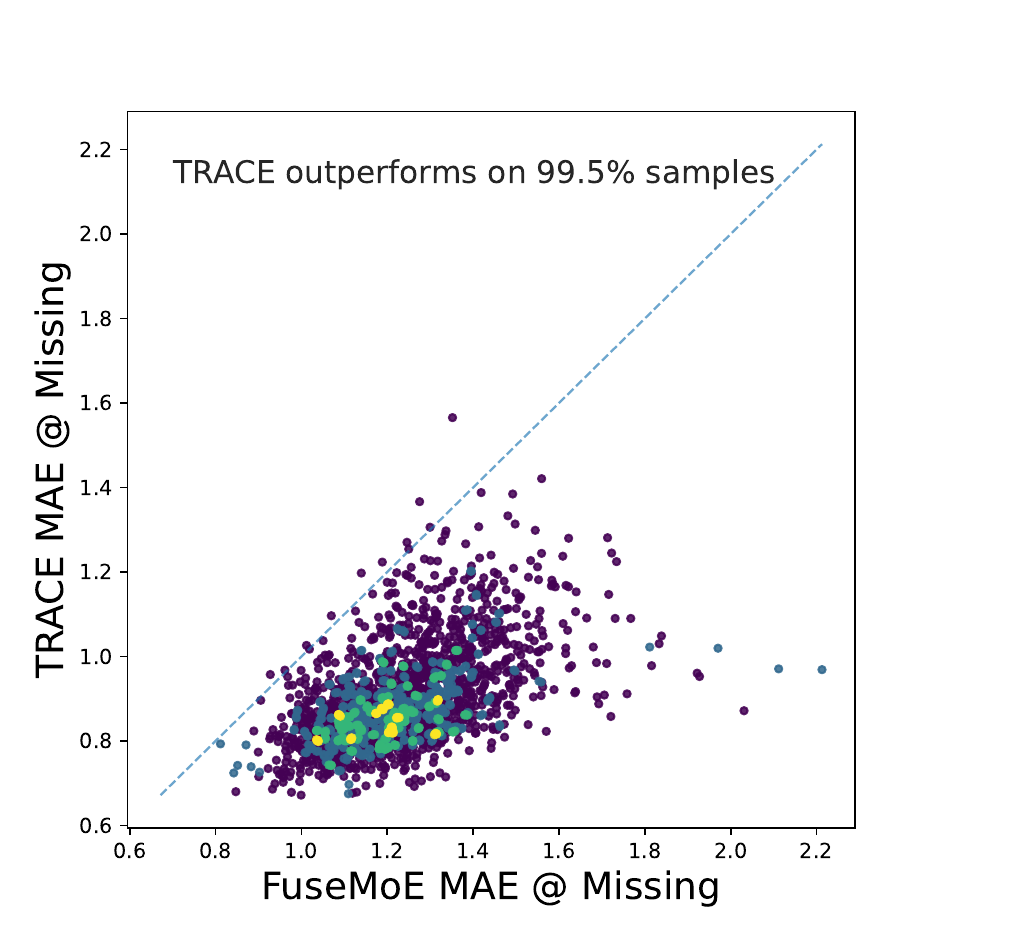} &
        \includegraphics[width=0.32\linewidth]{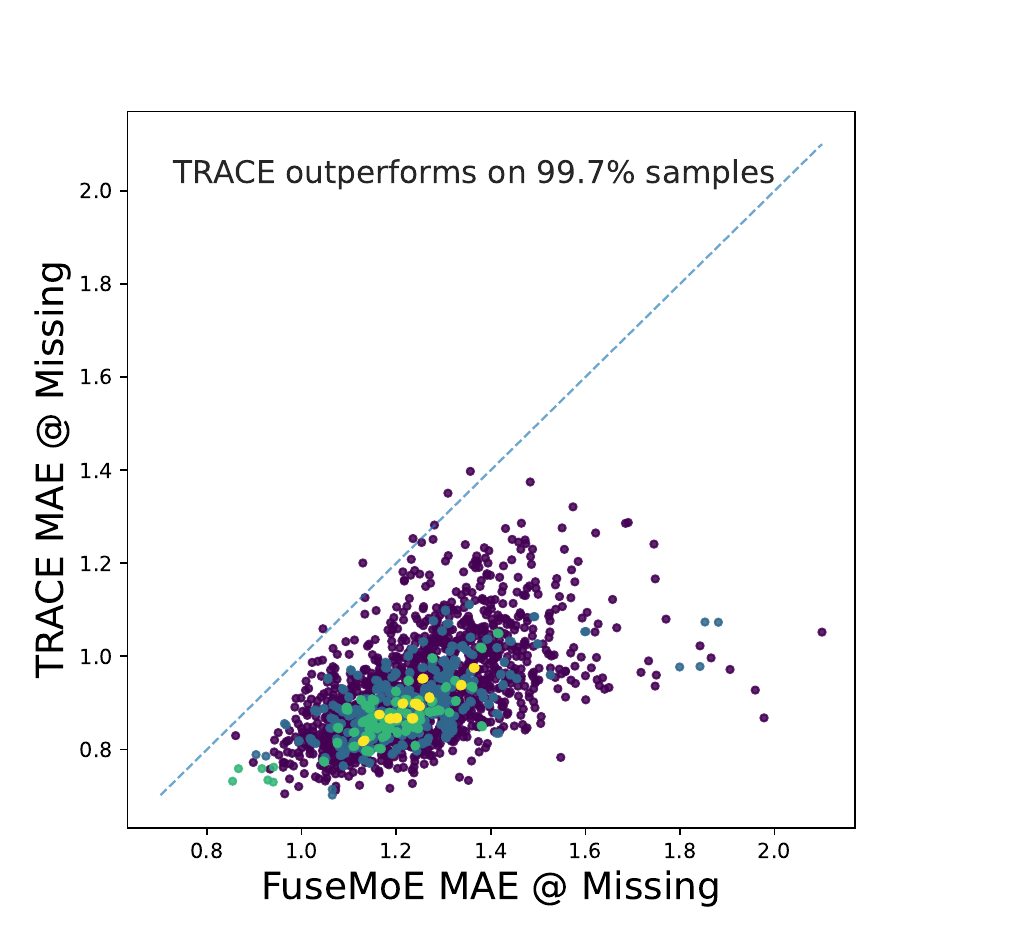} &
        \includegraphics[width=0.32\linewidth]{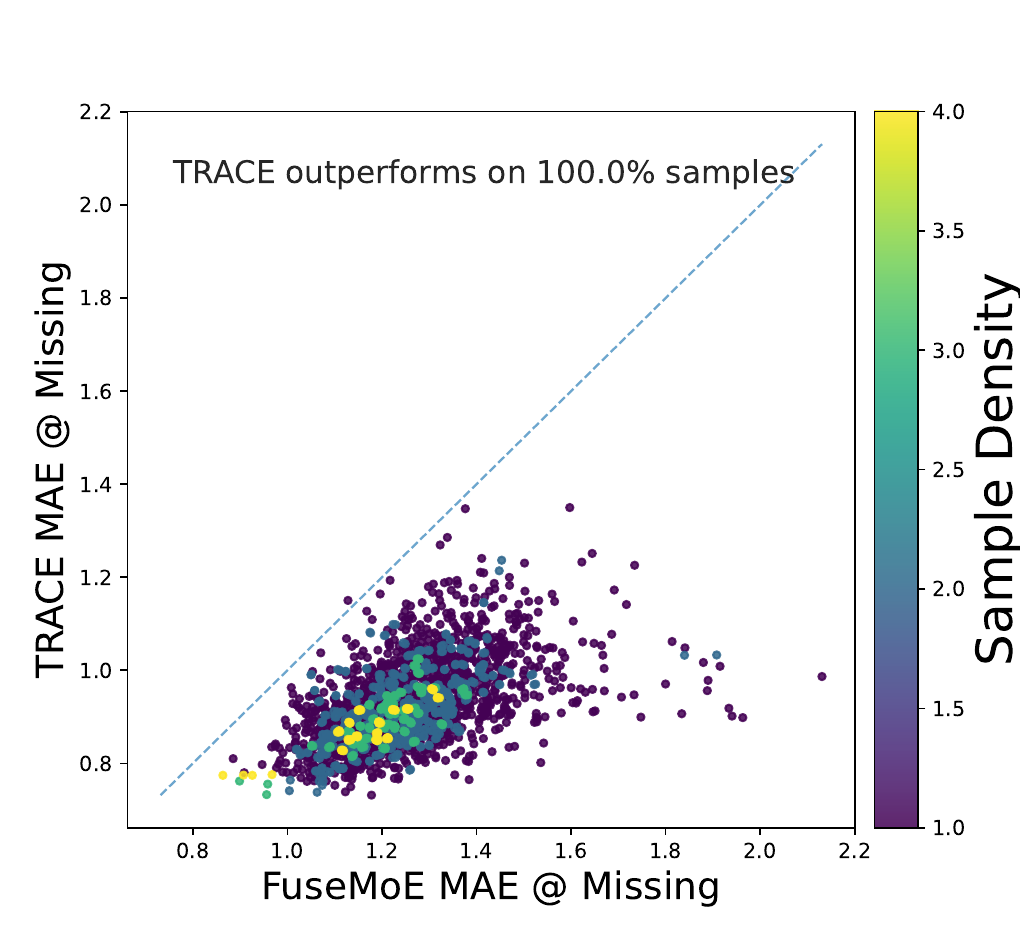} & \\
        \small MR=30\% & \small MR=40\% & \small MR=50\% 
    \end{tabular}
    \caption{
    \textbf{Motivational example under increasing missing rates.}
    Each panel compares per-sample MAE between FuseMoE and TRACE under a fixed missing rate (MR).
    As MR increases, the error distribution under TRACE increasingly concentrates below the parity line,
    indicating more consistent and robust conditional estimation in high-missing regimes.
    }
    \label{fig:additional_motivational_mr}

\end{figure*}

\begin{figure*}[t]
    \centering
    \setlength{\tabcolsep}{2pt}
    \begin{tabular}{cccc}
        \includegraphics[width=0.32\linewidth]{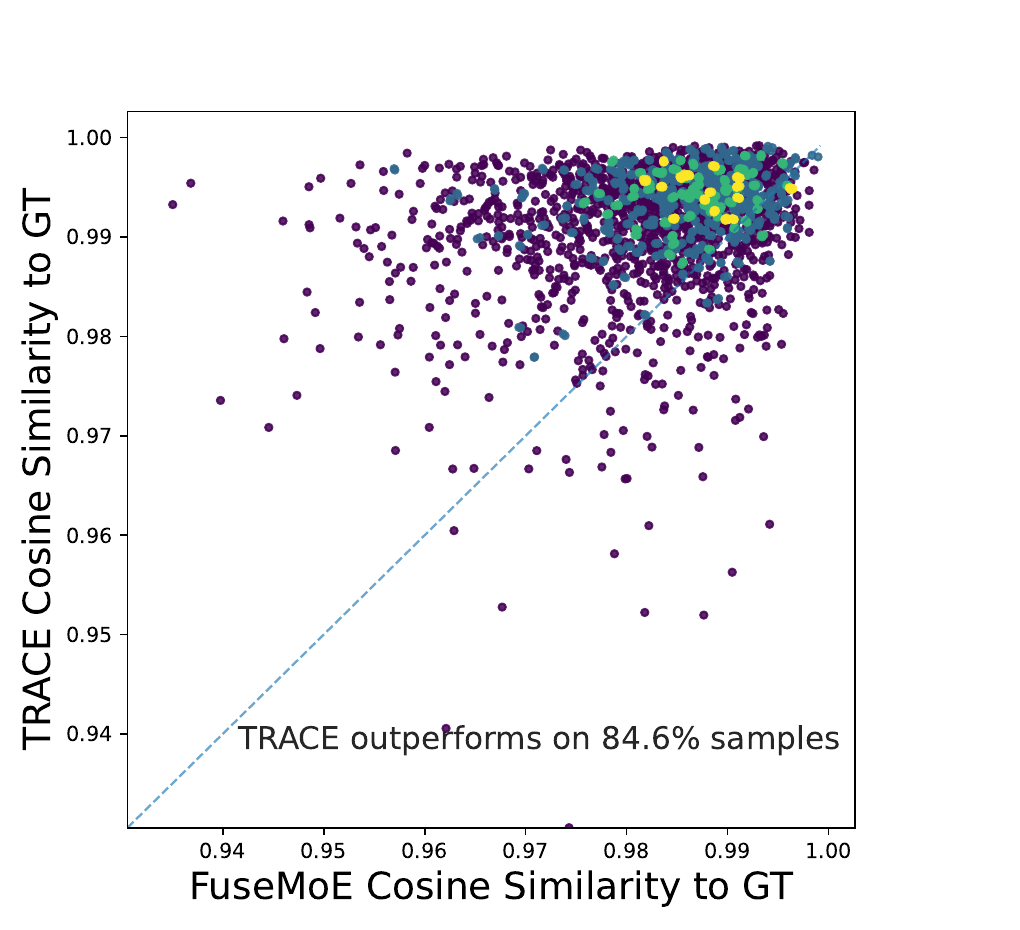}  &
        \includegraphics[width=0.32\linewidth]{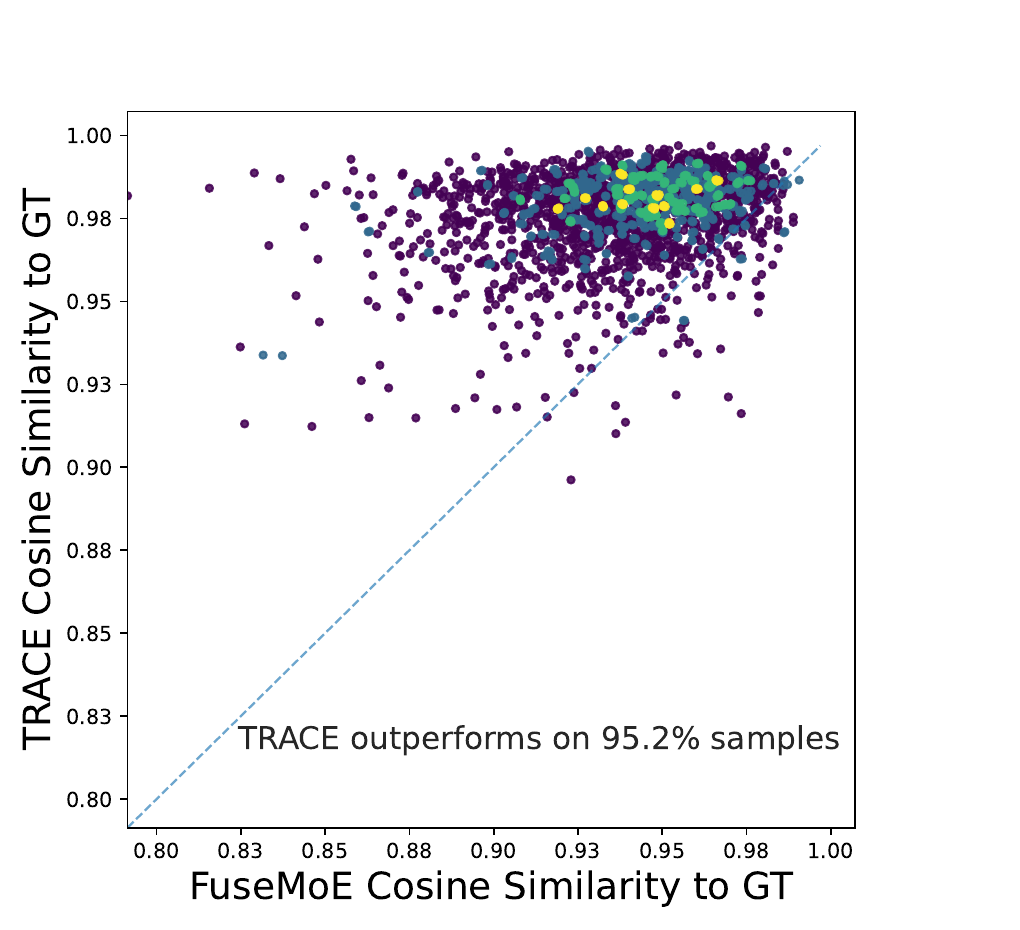}  &
        \includegraphics[width=0.32\linewidth]{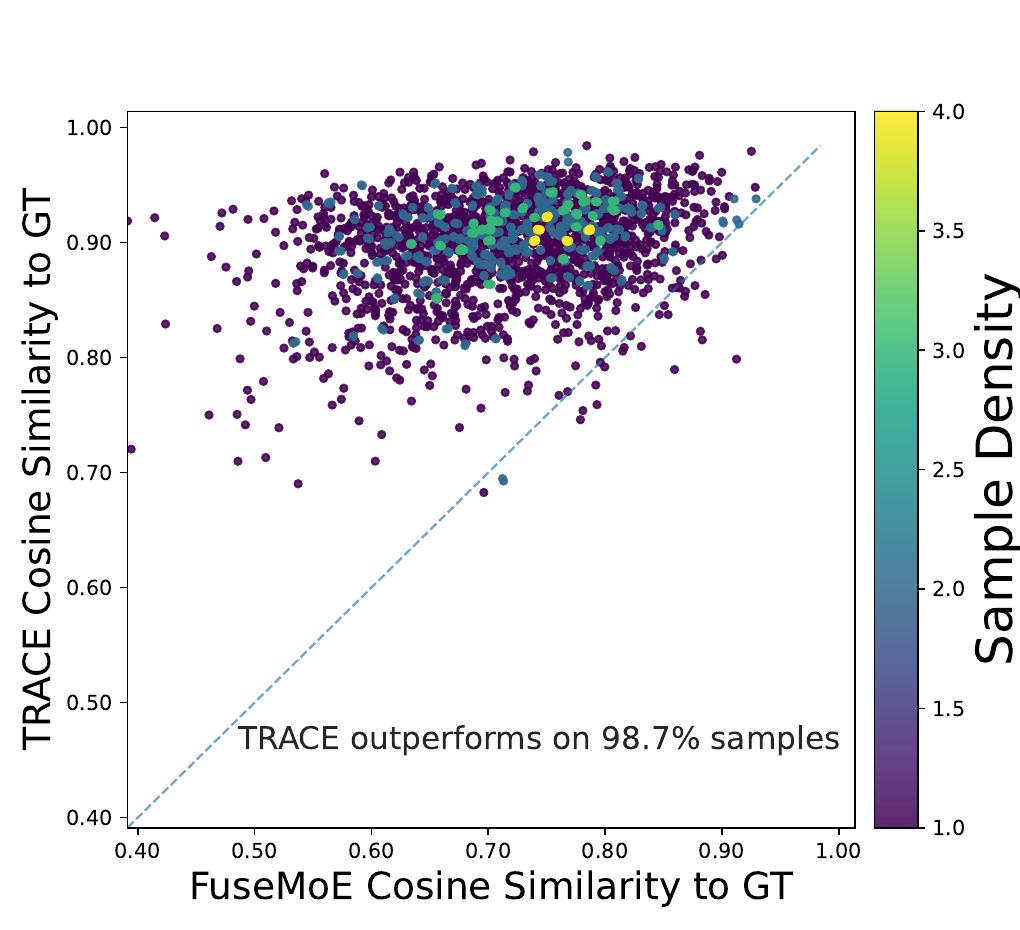} & \\
        \small MR=5\% & \small MR=10\% & \small MR=20\% \\[4pt]

        \includegraphics[width=0.32\linewidth]{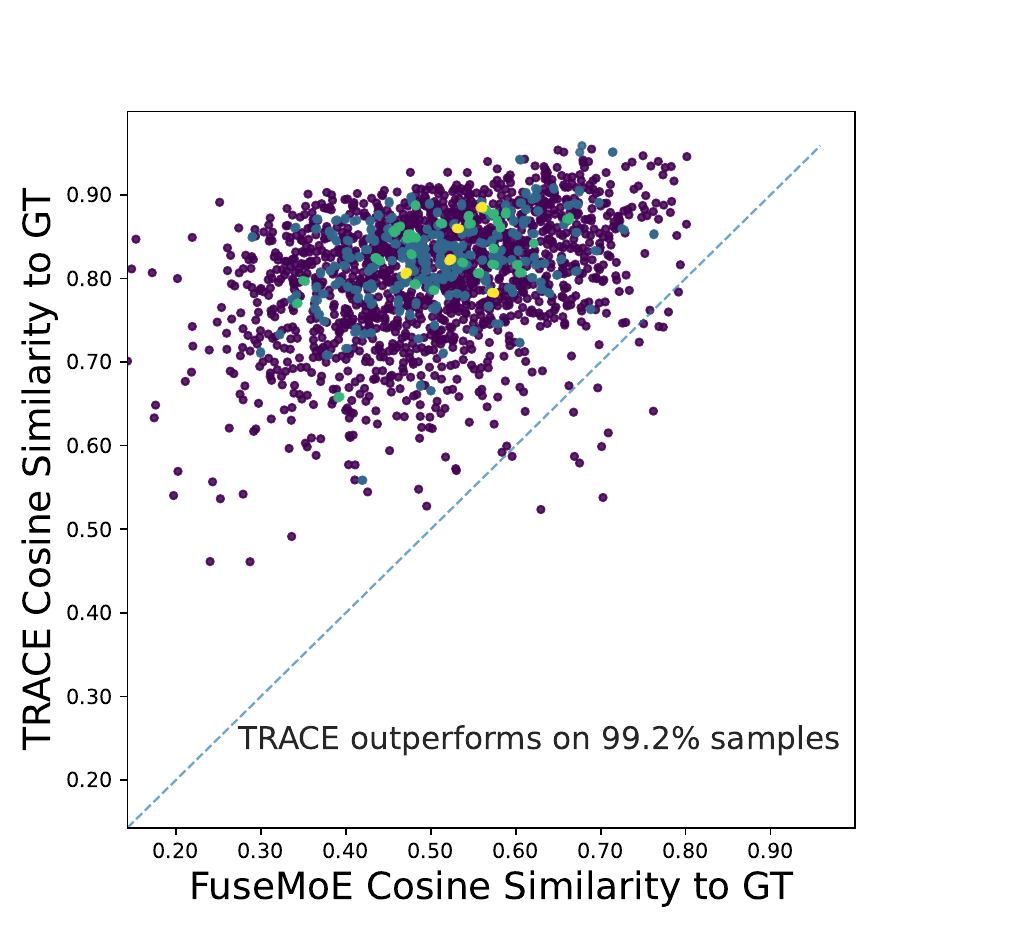} &
        \includegraphics[width=0.32\linewidth]{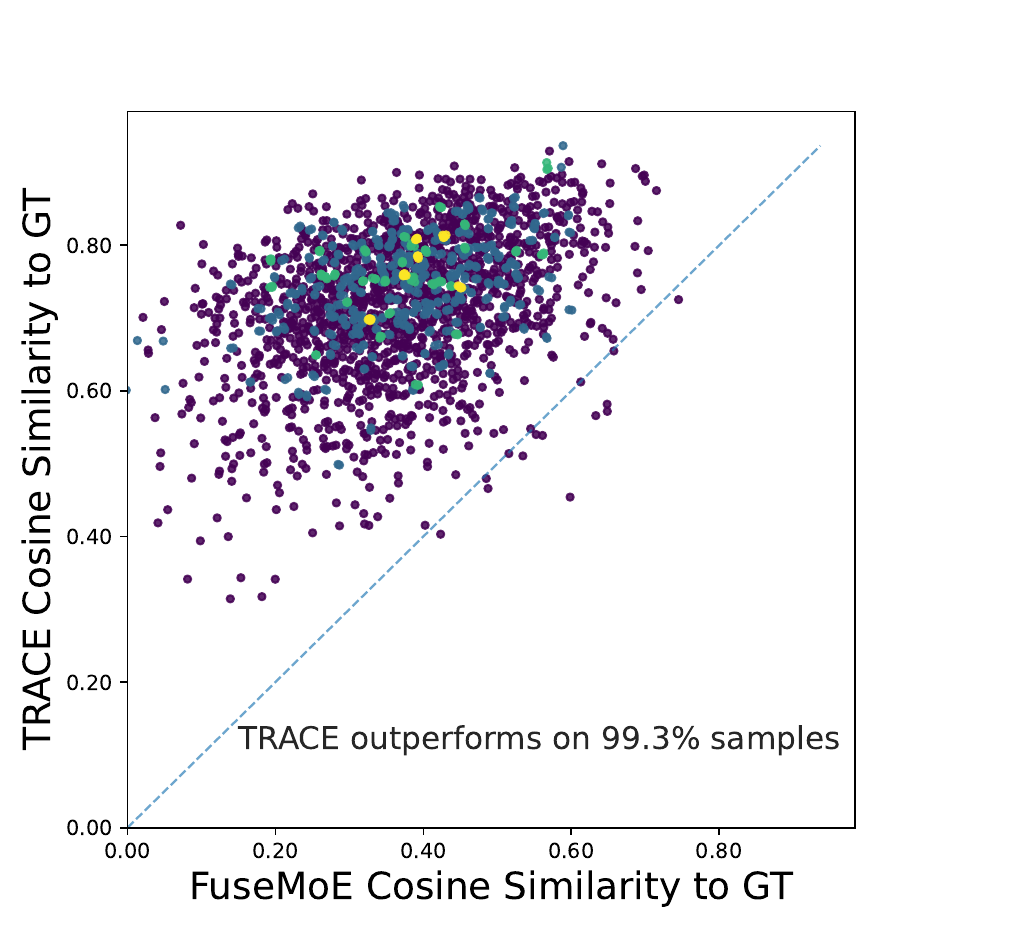} &
        \includegraphics[width=0.32\linewidth]{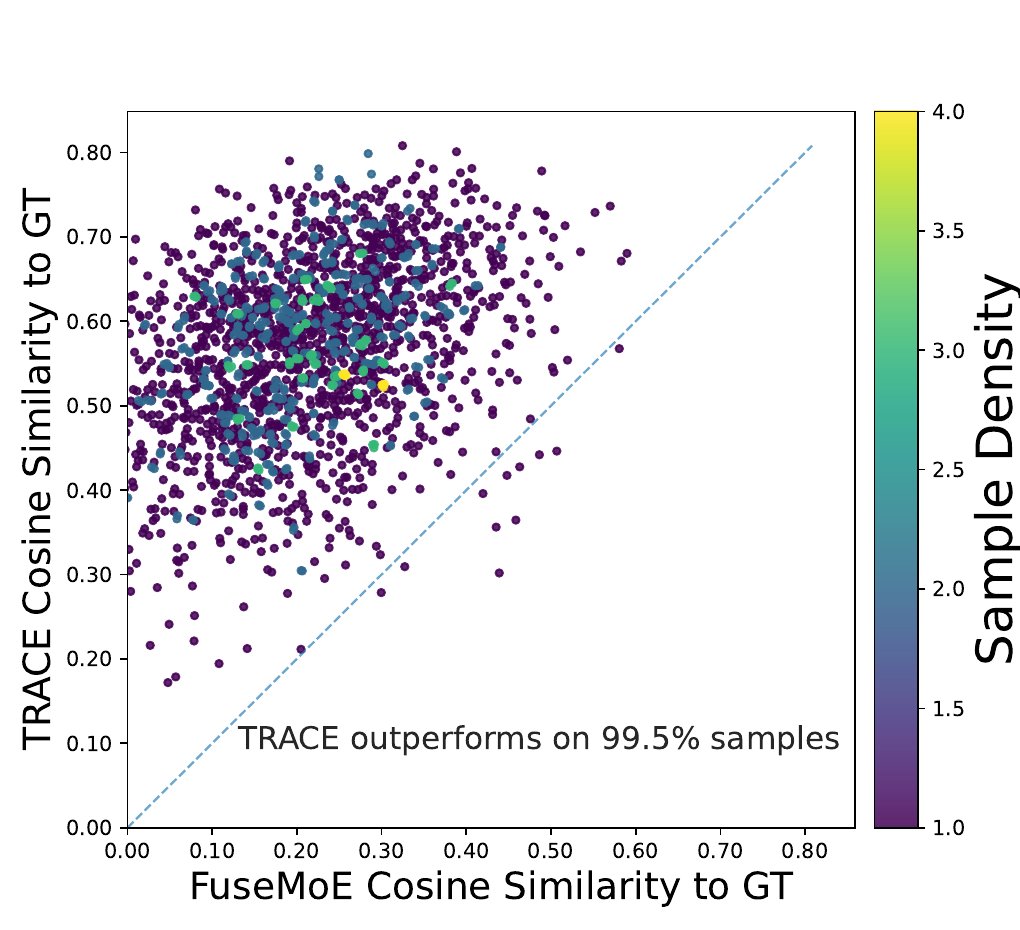} &\\
        \small MR=30\% & \small MR=40\% & \small MR=50\%
    \end{tabular}
    \caption{
    \textbf{Representation-level comparison under increasing missing rates.}
    Each panel shows the cosine distance between sequence-level representations
    obtained from imputed inputs and the corresponding oracle representations
    derived from fully observed inputs, under a fixed missing rate (MR).
    \label{fig:repr_motivational_mr}
    }
    
\end{figure*}

\subsection{Quantifying the Effect of Missing Rate on Imputation Quality}
\label{sec:missing_rate}
To systematically quantify how different imputation strategies behave under
increasing levels of modality missingness, we conduct a controlled comparison
that varies the missing rate (MR) from low to severe regimes.

Beyond measuring signal-level improvements, the primary goal of this analysis
is to examine whether the artifacts introduced by naive imputation can be fully
absorbed or corrected by subsequent fusion layers as observations become sparse,
or whether such artifacts persist and degrade the learned temporal
representations under high-missing regimes.

This evaluation is motivated by real-world clinical scenarios, where substantial
modality missingness is common rather than exceptional.
For instance, in the MIMIC-IV dataset, missing rates around 30\% are frequently
observed across multiple clinical variables, and certain modalities may exhibit
missingness exceeding 80\% due to irregular acquisition, clinical workflows, or
modality-specific availability.
Accordingly, we consider missing rates spanning from mild to high levels, covering
both well-observed and clinically realistic high-missing regimes.

\paragraph{Signal-level comparison.}
Figure~\ref{fig:additional_motivational_mr} reports a per-sample signal-level comparison
between FuseMoE and TRACE under progressively increasing missing rates.
When missingness is mild (MR $\leq$ 10\%), TRACE already outperforms FuseMoE on the
majority of samples, indicating a consistent but moderate advantage.
As the missing rate increases beyond moderate levels (MR $\geq$ 30\%), this gap
becomes increasingly pronounced, with TRACE achieving lower MAE@Missing on nearly
all samples ($>$ 99\%).
The error distributions under TRACE concentrate well below the parity line,
suggesting that naive imputation-based fusion degrades rapidly as direct
Observations become sparse.

\paragraph{Representation-level comparison.}
Figure~\ref{fig:repr_motivational_mr} presents a
representation-level analysis that compares sequence-level embeddings obtained
from imputed inputs against oracle representations derived from fully observed
inputs.
Across all missing rates, TRACE yields representations that are consistently
closer to the oracle in terms of cosine distance.
Notably, the representation gap between TRACE and FuseMoE widens as missingness
increases, exhibiting a trend that closely mirrors the signal-level behavior
observed in Figure~\ref{fig:additional_motivational_mr}.

Taken together, these results reveal a coherent pattern across both signal and
representation spaces.
As modality missingness increases, naive imputation-based fusion fails to preserve
the underlying temporal structure, and the artifacts introduced by naive imputation
are not fully absorbed or corrected by the subsequent fusion layers.
In contrast, conditional diffusion enables more robust signal reconstruction and
yields internal temporal representations that remain closer to the oracle under
severe missingness.
This analysis highlights that the advantage of TRACE is not limited to improved
pointwise imputation accuracy, but extends to preserving meaningful temporal
representations when direct observations become highly sparse.

\end{document}